\documentclass[]{fairmeta}

\usepackage[table]{xcolor}
\definecolor{lightblue}{rgb}{0.90, 0.95, 0.98}
\usepackage{xcolor}
\definecolor{customblue}{HTML}{C3D7F3}
\definecolor{softgreen}{HTML}{D9E7D5}
\definecolor{softpurple}{HTML}{BDAEE5}

\usepackage{multicol}
\usepackage{titletoc}
\usepackage{amsmath}  
\usepackage{bm}       
\usepackage{wrapfig}
\usepackage{xspace}
\usepackage{algorithm, algpseudocode}

\newcommand\method{MetaCanvas\xspace}
\newcommand{\thinparagraph}[1]{\vspace{0.15em}\noindent\textbf{#1}\enspace}

\definecolor{lightgreen}{rgb}{0.565, 0.933, 0.565}
\definecolor{lightpurple}{rgb}{0.784, 0.635, 0.784}

\usepackage{boldline}
\usepackage{array,multirow,graphicx}
\usepackage{enumitem}
\usepackage[dvipsnames]{xcolor}
\usepackage{amsmath, amssymb, amsfonts}

\title{Exploring MLLM-Diffusion Information Transfer with MetaCanvas}

\author[1,2,*]{Han Lin}
\author[1,3]{Xichen Pan}
\author[1,4,*]{Ziqi Huang}
\author[1]{Ji Hou}
\author[1]{Jialiang Wang}
\author[1,*]{Weifeng Chen}
\author[1,*]{Zecheng He}
\author[1]{Felix Juefei-Xu}
\author[1]{Junzhe Sun}
\author[1]{Zhipeng Fan}
\author[1]{Ali Thabet}
\author[2]{Mohit Bansal}
\author[1]{Chu Wang}

\affiliation[1]{Meta Superintelligence Labs}
\affiliation[2]{UNC Chapel Hill}
\affiliation[3]{New York University}
\affiliation[4]{Nanyang Technological University}

\contribution[*]{Work done at Meta}

\abstract{Multimodal learning has rapidly advanced visual understanding, largely via multimodal large language models (MLLMs) that use powerful LLMs as cognitive cores. In visual generation, however, these powerful core models are typically reduced to global text encoders for diffusion models, leaving most of their reasoning and planning ability unused. This creates a gap: current multimodal LLMs can parse complex layouts, attributes, and knowledge-intensive scenes, yet struggle to generate images or videos with equally precise and structured control. We propose \textbf{\method}, a lightweight framework that lets MLLMs reason and plan directly in spatial and spatiotemporal latent spaces and interface tightly with diffusion generators. 
We empirically implement \method{} on \emph{three} different diffusion backbones and evaluate it across \emph{six} tasks, including text-to-image generation, text/image-to-video generation, image/video editing, and in-context video generation, each requiring precise layouts, robust attribute binding, and reasoning-intensive control. \method{} consistently outperforms global-conditioning baselines, suggesting that treating MLLMs as latent-space planners is a promising direction for narrowing the gap between multimodal understanding and generation.
}

\date{\today}
\correspondence{\email{hanlincs@cs.unc.edu}, \email{wangchu@meta.com}}

\metadata[Website]{\url{https://metacanvas.github.io}}

\begin{document}

\maketitle

\section{Introduction}
\label{sec:intro}

Multimodal learning has primarily focused on multimodal understanding~\citep{liu2023visual, tong2024cambrian} and multimodal generation~\citep{stablediffusion, guo2023animatediff}. Recent progress in multimodal understanding typically leverages Large Language Models (LLMs)~\citep{touvron2023llama} as cognitive cores for building multimodal LLMs. These models achieve strong performance in visual question answering~\citep{antol2015vqa} and visual reasoning~\citep{johnson2017clevr}, enabled by visual instruction tuning~\citep{liu2023visual} that aligns them with perception encoders~\citep{Radford2021CLIP, siglip}. Using LLMs for multimodal understanding is highly effective because LLMs already excel at reasoning and question answering, and emerging evidence shows that they even acquire visual priors from language pretraining~\citep{han2025learning}.

However, in multimodal generation, LLMs are still largely treated as conditional encoders that map text prompts into representations for visual generative models~\citep{saharia2022imagen}, leaving most of their core capabilities underutilized. As a result, these models often struggle to produce visual content with accurate object positions and attributes~\citep{ghosh2023geneval}, or to handle cases requiring world-knowledge reasoning~\citep{wise}. In contrast, multimodal LLMs can easily interpret such structured visual content. This mismatch between the strengths of multimodal understanding and the limitations of multimodal generation has sparked interest in studying how these two abilities might complement each other~\citep{tong2024metamorph}, and how to more effectively transfer the rich capabilities of MLLMs into multimodal generative models.

Recent advances typically study this through the LLM+Diffusion paradigm, pairing pretrained (M)LLMs with external diffusion models that act as visual decoders~\citep{pan2025metaqueries, tong2024metamorph, liu2025step1x, lin2025bifrost, wu2025omnigen2, lin2025uniworld, koh2023GILL, ge2024seedx, chen2025blip3ofamilyfullyopen, pankosmos, yin2025best}. In these two-component systems, the (M)LLMs first processes all contextual signals that guide generation, and their outputs are then injected into a separate diffusion generator via attention. Specifically, MetaQuery~\citep{pan2025metaqueries} provides early evidence that learnable query embeddings from frozen LLMs can serve as effective conditions, with improved performance in transferring LLM knowledge and in-context learning capabilities to reasoning-augmented image generation.

The optimal interface between (M)LLMs and diffusion models remains unclear. Existing approaches typically use the output of (M)LLMs as a global condition for visual generation~\citep{pan2025metaqueries, lin2025uniworld, wu2025omnigen2}. Although this provides useful guidance to the diffusion decoder, it constrains the ability of (M)LLMs to deliver fine-grained spatial or temporal controls throughout the diffusion process. In practice, the control required for visual generation is inherently structured and region-specific~\citep{zhang2023controlnet}. Global conditions alone struggles to compel (M)LLMs to explicitly specify where objects should appear, how spatial relations should be grounded, or how temporal dependencies should evolve. In this paper, we wish to investigate whether (M)LLMs can instead reason and plan directly within spatial and spatiotemporal latent spaces, and whether these capabilities would benefit multimodal generation. To do so, we explore an alternative approach: enforcing explicit, patch-by-patch control over the generation process using a learnable latent canvas whose layout is planned by the (M)LLM.

\begin{figure*}[t!]
  \centering
  \includegraphics[width=\linewidth]{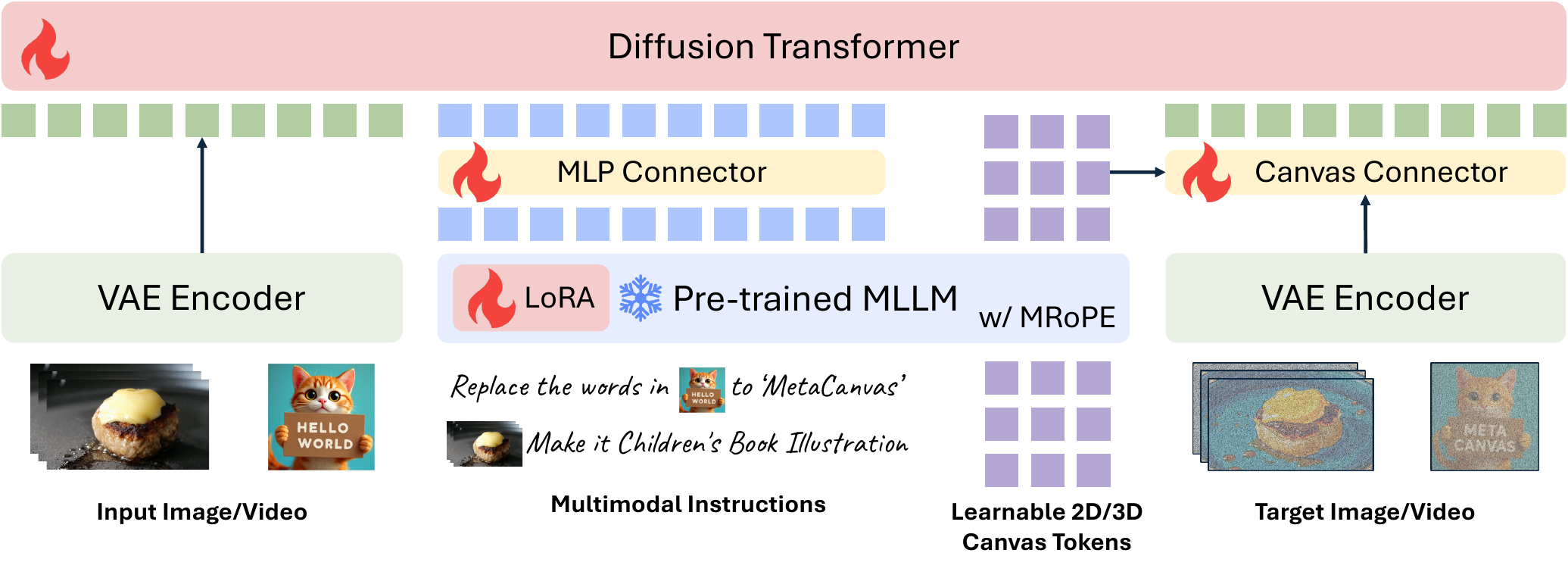}
  \vspace{-8mm}
  \caption{
  \textbf{Overview of the \method{} framework.}
  \method{} tokenizes the text and encodes it using the MLLM’s text embedder, while user-provided images and videos are encoded using both the MLLM’s visual encoder and the VAE encoder. The text embeddings produced by the MLLM are passed through a lightweight MLP connector and used as conditioning for the DiT. In addition, we append a set of learnable multidimensional canvas tokens to the MLLM input, which are processed using multimodal RoPE~\citep{Qwen2.5-VL}. The resulting canvas embeddings are then fused with the noisy latents through a lightweight transformer-based connector with two blocks. Connector details are illustrated in~\Cref{fig:connector_design}.
  \colorbox{softgreen}{Green} tokens represent media context tokens,
  \colorbox{customblue}{blue} tokens represent text context tokens, and \colorbox{softpurple}{purple} tokens represent the {canvas} tokens.
  } 
  \vspace{-3mm}
\label{fig:overview_model_architecture} 
\end{figure*}

\begin{figure}[t]
  \centering
  \includegraphics[width=0.55\linewidth]{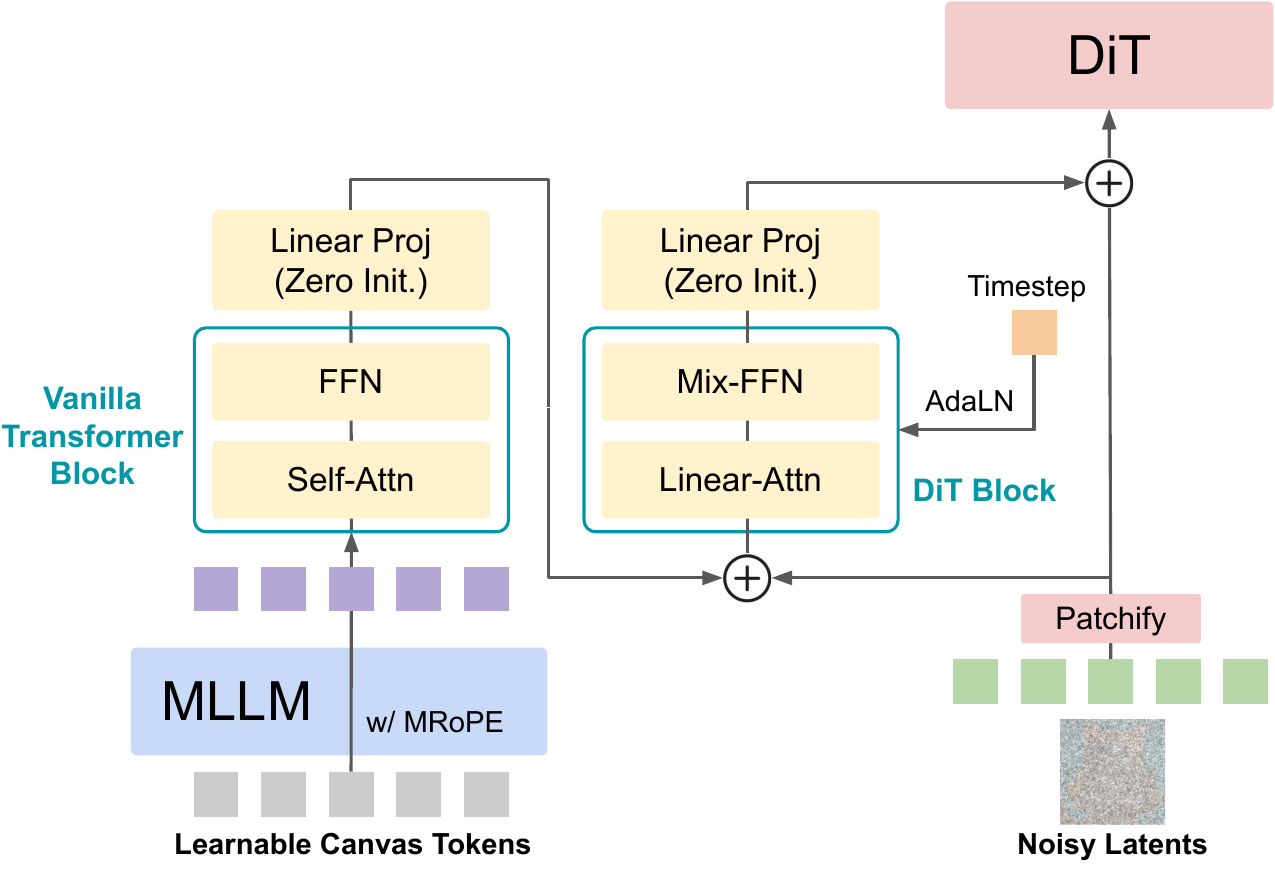}
  \vspace{-2mm}
  \caption{
  \textbf{\method connector design details.}
  The connector comprises a vanilla Transformer block and a Diffusion Transformer (DiT) block. The vanilla Transformer block transforms the learnable canvas tokens to align them with the DiT latent space. The second DiT block adopts a ControlNet-style design, where the transformed canvas tokens and the noisy latents are first combined and then passed through a DiT block with Adaptive LayerNorm~\citep{perez2018film}. We adopt Linear-Attn and Mix-FFN design from~\citep{xie2024sana} to reduce memory usage. The outputs of both blocks are followed by a zero-initialized linear projection layer, ensuring that at the beginning of training, the learnable canvas tokens have no influence on the DiT’s latent inputs.
  }
    \vspace{-2mm}
\label{fig:connector_design} 
\end{figure}

Concretely, we present \textbf{\method}, a lightweight architecture that connects pretrained MLLMs with diffusion-based visual generators (see~\Cref{fig:overview_model_architecture} for an overview). The key idea is to have the MLLM produce implicit visual sketches of the desired output, serving as spatial or spatiotemporal priors for guiding the diffusion process. Specifically, \method appends a set of learnable, multidimensional canvas tokens to the MLLM’s multimodal input sequence and processes them using multimodal RoPE~\citep{Qwen2.5-VL}. The MLLM outputs embeddings for these canvas tokens, which are then passed into a lightweight connector module (see~\Cref{fig:connector_design} and ~\Cref{subsec:method_connector_design} for design details). This connector comprises two Transformer blocks that align the canvas tokens and inject them into the diffusion model’s noisy latents patch by patch.
To stabilize training, we follow the zero-initialization strategy from~\citep{zhang2023controlnet}, ensuring the injected signals initially leave the diffusion model’s inputs unchanged.

We demonstrate the effectiveness of \method{} through extensive experiments. First, in \Cref{subsec:results_exploratory_experiments_t2i}, we conduct an exploratory small-scale text-to-image (T2I) generation experiment. We visually validate that canvas tokens extracted from the MLLM can serve as reasonable visual planners to guide image synthesis in DiT (see~\Cref{fig:attention_map_visualization}), and quantitatively observe that \method{} converges faster and consistently outperforms other design variants on the GenEval~\citep{ghosh2023geneval} benchmark.
Motivated by this positive signal, in \Cref{subsec:results_image_editing}, we extend \method to image editing tasks and scale up the architectures to Qwen2.5-VL-7B~\citep{Qwen2.5-VL} paired with FLUX.1-Kontext-Dev~\citep{batifol2025flux}, and observe consistent gains in both training efficiency and editing benchmark performance (see~\Cref{table:results_gedit_short} and~\Cref{table:results_imgedit_short}). 
To evaluate the generalizability of \method{} across tasks, we extend it to video generation and editing in \Cref{subsec:results_video_generation_editing}. By adopting a multi-dimensional canvas design (\Cref{subsec:method_multidimensional_canvas_tokens}), we integrate Qwen2.5-VL-7B with Wan2.2-5B~\citep{wan2025wan} through a multi-stage training procedure. This design unifies diverse tasks, including text-to-video (T2V), image-to-video (I2V), and video editing (V2V), and naturally enables new capabilities such as reference-guided video generation, allowing flexible multimodal inputs (e.g., text combined with reference images or videos) while fully preserving the MLLM’s core multimodal understanding. Notably, while unlocking multiple tasks, \method preserves the original video generation capabilities (see~\Cref{table:results_t2v_i2v}) and achieves substantial improvements in V2V prompt accuracy and overall video quality compared to the latest open-source methods~\citep{decart2025lucyedit,wu2025insvie,bai2025scaling}, attaining the highest overall evaluation scores in both VLM-based and human assessments (see~\Cref{table:results_video_editing}). Finally, we also demonstrate that \method{} generalizes to in-context video generation tasks, achieving competitive or superior performance compared to previous methods (see~\Cref{table:results_omnicontext_video}).

In short, our contribution can be summarized as follows:
\begin{itemize}[leftmargin=1em]
    \setlength{\itemsep}{0pt}
    \setlength{\parskip}{0pt}
    \setlength{\parsep}{0pt}
    \item \textbf{Latent canvas connector for MLLM-to-diffusion.}
    \method{} uses learnable canvas tokens as \emph{implicit visual sketches} and a lightweight connector to inject them patch-wise into diffusion latents, providing a simple and effective interface.

    \item \textbf{Spatial–temporal planning in latent space.}
    \method{}’s multidimensional canvas queries enable the MLLM to plan object layouts and temporal evolution directly in latent space, improving structural fidelity and providing fine-grained editing control.
    
    \item \textbf{Performance and generalizability.}
    \method{} generalizes across six diverse image and video tasks and three diffusion backbones (i.e., MMDiT and cross-attention). It surpasses baseline without canvas tokens, and matches state-of-the-art performance while preserving the MLLM’s multimodal understanding.
\end{itemize}

\section{Related Work}
\label{sec:related_work}

\thinparagraph{Text as the interface to generators.}
The most direct way to connect LLMs with visual generators is to use the LLM’s text output as the interface: the model expands a prompt into detailed captions, layouts, or scripts that condition a diffusion model~\citep{li2023gligen, yang2023reco, feng2023layoutgpt, Cho2023VPT2I, linvideodirectorgpt, zaladiagrammergpt, lianllm, lian2023lmd, huang2025vchain, qu2023layoutllm}. However, text-only signals provide limited guidance, they cannot reliably encode dense layouts, precise attribute bindings, or long-range temporal structure required for complex, compositional scenes. A more expressive alternative is to condition generation using dense embeddings from MLLMs.

\thinparagraph{Embeddings from (M)LLMs as the interface.}
To increase the bandwidth between (M)LLMs and diffusion backbones, recent methods directly feed dense embeddings from (M)LLMs into the generator. The (M)LLM processes text, images, or videos and either autoregressively predicts visual representations~\citep{sun2023emu, tong2024metamorph} or encodes query tokens into continuous embeddings~\citep{ge2024seedx, dongdreamllm, pan2025metaqueries, chen2025blip3ofamilyfullyopen}. This interface is more expressive than pure text and largely preserves the (M)LLM’s reasoning ability, but it still compresses geometry, layout, and motion into a single 1D sequence. Because these embeddings are used only as a global, sequence-level condition, the diffusion model lacks explicit spatial or temporal handles for generation. This motivates introducing a structured spatial or spatiotemporal latent canvas into which the (M)LLM can directly write patch-wise plans for the diffusion generator to follow.

\section{\method}
\label{sec:method}

We introduce \textbf{\method}, a novel unified multimodal learning framework that effectively bridges pretrained MLLMs and diffusion models with learnable multi-dimensional canvas tokens and a lightweight connector, which improves information transfer between MLLMs and diffusion models with efficient training.
We first introduce preliminaries about multimodal conditioning via MLLMs in \Cref{subsec:method_preliminaries}, then discuss the overall framework in \Cref{subsec:method_overall_framework}, followed by canvas tokens design in \Cref{subsec:method_multidimensional_canvas_tokens} and connector design in \Cref{subsec:method_connector_design}.

\subsection{Preliminaries}
\label{subsec:method_preliminaries}

\thinparagraph{Multimodal conditioning via MLLM.}
In bridging methods that connect MLLMs and diffusion models, the text conditioning tokens $\bm{c}_{\text{text}}$ in the latent diffusion model are replaced by the output tokens from the MLLM. Specifically, the input modalities including text, {images and videos are first encoded by the matching tokenizer}, 
then passed through the MLLM backbone. The resulting text (and visual) tokens are used as conditions to guide the diffusion model.
In \method, we further append a set of learnable canvas tokens to the MLLM's multimodal input sequence and process them jointly. The output embeddings 
of these canvas tokens are used as structural {visual-semantic} priors and are added to the noisy latents $\bm{z}_t$ to provide additional and direct {diffusion canvas} guidance during generation.

\subsection{Overall Framework for \method}
\label{subsec:method_overall_framework}

We illustrate the overall \method framework in~\Cref{fig:overview_model_architecture}. Given user instructions with text and visual inputs, \method tokenizes the text and encodes it with the MLLM’s text embedder, while images or videos are encoded by the MLLM’s visual encoder and also by the diffusion model’s VAE encoder for both semantic and low-level details.

\thinparagraph{{Context} token conditioning.}
Embeddings of the multimodal input tokens produced by the MLLM (i.e., context tokens for generation) are passed through a lightweight two-layer MLP connector and given to the DiT via cross-attention or self-attention, depending on the DiT’s architecture, following standard practice in~\citep{pan2025metaqueries, tong2024metamorph, liu2025step1x, lin2025bifrost, lin2025uniworld, ge2024seedx, chen2025blip3ofamilyfullyopen}. Further discussion is provided in \Cref{subsec:appendix_context_canvas_connectors_design}.

\thinparagraph{{Canvas} token conditioning.}
We append a set of {\sl learnable} multidimensional canvas tokens to the MLLM input, which are processed using multimodal RoPE~\citep{Qwen2.5-VL}. The resulting canvas embeddings are then fused additively with the noisy latents through a lightweight \emph{Canvas Connector} (see~\Cref{subsec:method_connector_design}), ensuring strong alignment between the MLLM-drafted canvas and the actual generation latents.

\thinparagraph{Training and inference.}
The main learnable components of \method{} include the MLP connector, the canvas connector, and the DiT (LoRA-finetuned in~\Cref{subsec:results_exploratory_experiments_t2i}, finetuned only on the visual branch in~\Cref{subsec:results_image_editing}, or fully finetuned in~\Cref{subsec:results_video_generation_editing}). During training, the MLLM is kept frozen, {\sl optionally} with an additional LoRA to enhance its visual generation capacity (ablations in \Cref{subsec:appendix_additional_ablation_results}). During inference, canvas tokens are appended to the MLLM only after the end-of-sentence \texttt{<EoS>} token. If a LoRA is enabled, it remains inactive until \texttt{<EoS>} and is activated only after the canvas tokens are appended. This design preserves the MLLM’s understanding and reasoning capabilities while enabling strong generative and steering performance through lightweight, low-cost trainable components.

\begin{figure}[t!]
  \centering
  \includegraphics[width=0.55\linewidth]{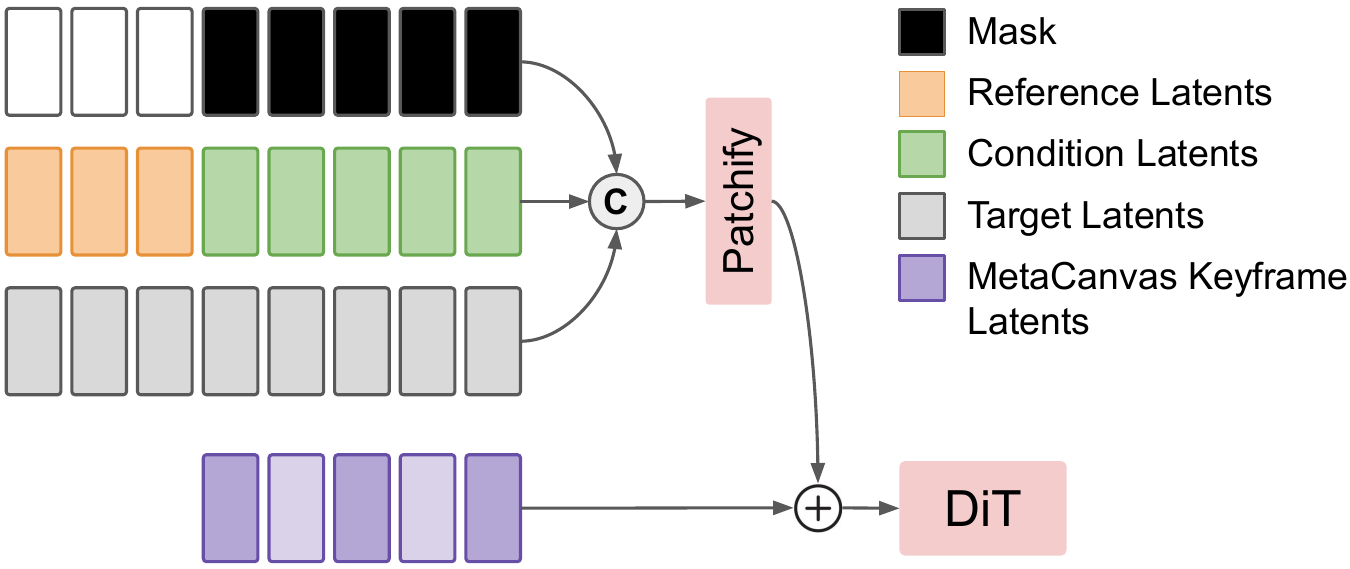}
  \vspace{-1mm}
  \caption{
  \textbf{\method keyframes and reference/condition frames injection strategy for video tasks.} We modify the input layer of Wan2.2-5B~\citep{wan2025wan} to concatenate reference and condition latents with noisy latents along the channel dimension. The resulting tensor is then passed through the patch embedding layer and combined with \method keyframes after interpolation. Light purple tokens represent interpolated keyframe canvas. Note that we do not apply \method keyframe latents to reference frames for video tasks.  
  }
  \vspace{-2mm}
\label{fig:video_patch_embed_design} 
\end{figure}

\subsection{Multi-dimensional Canvas Tokens}
\label{subsec:method_multidimensional_canvas_tokens}

\textbf{Keyframe canvas design.} For image generation and editing tasks,
\method{} employs 2D canvas tokens whose spatial arrangement adapts to the image resolution. For videos, instead of using a dense 3D canvas covering all latent frames, we introduce learnable sparse keyframe canvas tokens that capture representative temporal information and are then \textit{linearly interpolated to the full latent-frame space} before being incorporated into the DiT latents.
This strategy preserves temporal coherence while keeping the token interface compact and efficient for both training and inference. Additional details regarding the image and video canvas design are provided in \Cref{subsec:appendix_2d_vs_3d_canvas}.

\thinparagraph{Input layer adaptation for video tasks.} As illustrated in~\Cref{fig:video_patch_embed_design}, and following~\citep{cheng2025wan}, we modify the input layer of Wan2.2-5B~\citep{wan2025wan} to concatenate the reference and condition latents with the noisy latents along the channel dimension. These latents are then passed through the patch embedding layer and combined with the \method{} keyframe tokens after interpolation (detailed below).

\subsection{\method Connector Design}
\label{subsec:method_connector_design}

The connector for learnable canvas tokens is {designed} with the following components. Ablation {studies} on the effectiveness of these components are shown in~\Cref{table:geneval_method_ablations}.
\begin{itemize}[leftmargin=1em]
    \setlength{\itemsep}{0pt}
    \setlength{\parskip}{0pt}
    \setlength{\parsep}{0pt}
    \item \textbf{Vanilla Transformer block.} We first pass the learnable canvas tokens, being 2D canvas for image case or 3D keyframe canvas for video case, through a single Transformer block to {transfer and align} their features with the DiT latent space. Note that for video tasks (\Cref{subsec:method_multidimensional_canvas_tokens}), we will additionally perform linear interpolation of the output tokens from this block to match the noisy latents shape of the DiT.
    \item \textbf{DiT block.} Next, we add the canvas tokens to the DiT noisy latents after the patchify layer. For video tasks, as shown in \Cref{fig:video_patch_embed_design}, we {only} add canvas tokens to the noisy latent frames while avoiding any modification of the reference frames. The fused canvas tokens are then processed using an efficient linear-attention DiT block in SANA~\citep{xie2024sana} and ultimately added back to the original DiT latent space. We adopt AdaLN~\citep{perez2018film} in the connector design to dynamically modulate the influence of the canvas tokens across different timesteps.
    \item \textbf{Zero-initialized linear projections.} After each Transformer block, we apply a zero-initialized linear projection, inspired by~\citep{zhang2023controlnet}, to ensure that at the start of training the canvas branch produces a null residual and does not modify the DiT latent inputs.
    \item \textbf{Patchify then fuse.} We find that fusing canvas tokens with the noisy latents after patchification yields better performance, as it avoids projecting high-dimensional canvas tokens into the lower-dimensional VAE space, which would otherwise result in information loss.
\end{itemize}

\section{Experiment Setup}
\label{sec:exp_setup}

\subsection{Training Setup}
\label{subsec:exp_setup_training}

\thinparagraph{Exploratory experiment on T2I generation task.}
To quickly test \method and assess the connector design, we begin with exploratory experiments using Qwen2.5-VL-3B~\citep{Qwen2.5-VL} and SANA-1.6B~\citep{xie2024sana}. We retain SANA’s original T5 text encoder for text conditioning and add additional token conditions using different methods from Qwen2.5-VL-3B to evaluate information transfer from the MLLM to the DiT. For canvas tokens, we adopt the transformer-based connector described in~\Cref{subsec:method_connector_design}. For MetaQuery-style 1D query tokens, we reuse the text-conditioning interface and concatenate the MLLM’s query tokens with the T5 text tokens as a joint condition to the DiT. We keep the MLLM frozen, and train the connectors, canvas/query tokens, as well as the DiT {\sl from scratch}. We use BLIP3o-60k~\citep{chen2025blip3ofamilyfullyopen} as the training data. Training hyperparameters are listed in \Cref{sec:appendix_exploratory_t2i_generation}.

\thinparagraph{Image editing task.}
As FLUX.1-Dev~\citep{flux} and FLUX.1-Kontext-Dev~\citep{batifol2025flux} share the same T5 text encoder, we follow GPT-Image-Edit~\citep{wang2025gpt} and initialize from the MLP connector in UniWorld-V1~\citep{lin2025uniworld}, which was trained to bridge Qwen2.5-VL-7B~\citep{Qwen2.5-VL} and FLUX.1 [Dev]~\citep{flux}. We start the training directly by unfreezing the diffusion model’s visual branch, together with the additional learnable canvas tokens and the transformer-based connector. To increase MLLM's model capacity, we apply LoRA with rank of 64 to the MLLM backbone. 
The model is trained on $\mathcal{O}$(1M)  image-editing samples. See \Cref{sec:appendix_image_editing_task} for training details.

\thinparagraph{Video generation and editing tasks.}
Wan2.2-5B~\citep{wan2025wan} uses T5 text encoder by default. To better support multimodal inputs, we replace T5 with a MLLM (Qwen2.5-VL-7B~\citep{Qwen2.5-VL}) rather than fusing MLLM and T5. Hybrid MLLM+T5 features can introduce conflicting optimization signals~\citep{lin2025uniworld}, and early reliance on T5 often drives training into poor local minima, whereas a single MLLM pathway yields more stable convergence and better alignment.
We employ a three-stage training strategy (see \Cref{sec:appendix_video_tasks_training_details} for details): 
\begin{itemize}[leftmargin=1em]
    \setlength{\itemsep}{0pt}
    \setlength{\parskip}{0pt}
    \setlength{\parsep}{0pt}
    \item \textbf{Stage 1: Connector alignment.} In the first stage, we use $\mathcal{O}$(40M) image-text pairs mainly focusing on image-only training, with the goal of aligning the image semantics between the two models. As we train on relatively low resolution images, we freeze the whole MLLM and diffusion model, and \textit{only train the connector}.
    \item \textbf{Stage 2: High-resolution finetuning.} In the second stage, we {further} incorporate $\mathcal{O}$(3M) video-text pairs, and set the video:image data ratio as 4:1, with standard resolutions and video length of 121 frames, to let the model learn motions. We \textit{unfreeze the cross-attention layers} in diffusion models in this stage for better alignment.
    \item \textbf{Stage 3: Multitask training.} In the third stage, we train on diverse image and video tasks, \textit{unfreeze all parameters in diffusion models, and add LoRA on MLLM as well as trainable parameters in \method connectors}. We train the model jointly on diverse image and video tasks. Task-specific dataset details are illustrated in the Appendix.
\end{itemize}

\subsection{Evaluation Setup}
\label{subsec:exp_setup_evaluation}
We evaluate \method{} on a wide variety of benchmarks:
\begin{itemize}[leftmargin=1em]
    \setlength{\itemsep}{0pt}
    \setlength{\parskip}{0pt}
    \setlength{\parsep}{0pt}
    \item \textbf{Exploratory experoments on T2I generation:}
Evaluate attributes (count, color, spatial, etc.) on GenEval~\citep{ghosh2023geneval}.
    \item \textbf{Image editing: }
Following prior works~\citep{lin2025uniworld, wu2025qwen, deng2025emerging, wu2025omnigen2}, we report results on GEdit~\citep{liu2025step1x} and ImgEdit~\citep{ye2025imgedit}, using GPT-4o~\citep{openai_gpt4o_2024} as the evaluator.
    \item \textbf{Video generation:}
For T2V and I2V generation tasks, we use VBench~\citep{huang2023vbench, huang2024vbench++} as the evaluation benchmark.
    \item \textbf{Video editing:}
For video editing task, due to the scarcity of open-source video editing evaluation benchmarks with high resolution (720p) and sufficient duration (121 frames at 24 FPS), we curate a balanced evaluation set consisting of 300 video prompts. Dataset details are illustrated in \Cref{sec:appendix_video_editing_task}.
    \item \textbf{In-context video generation:}
We evaluate reference image-to-video generation, introducing \textit{OmniContext-Video} by extending the image-generation benchmark OmniContext~\citep{wu2025omnigen2} to video, covering both single-ID and multi-ID scenarios across objects, characters, and scenes. Dataset details are illustrated in \Cref{sec:appendix_in_context_video_generation_task}.
\end{itemize}

\begin{figure}[t]
  \centering
  \includegraphics[width=0.65\linewidth]{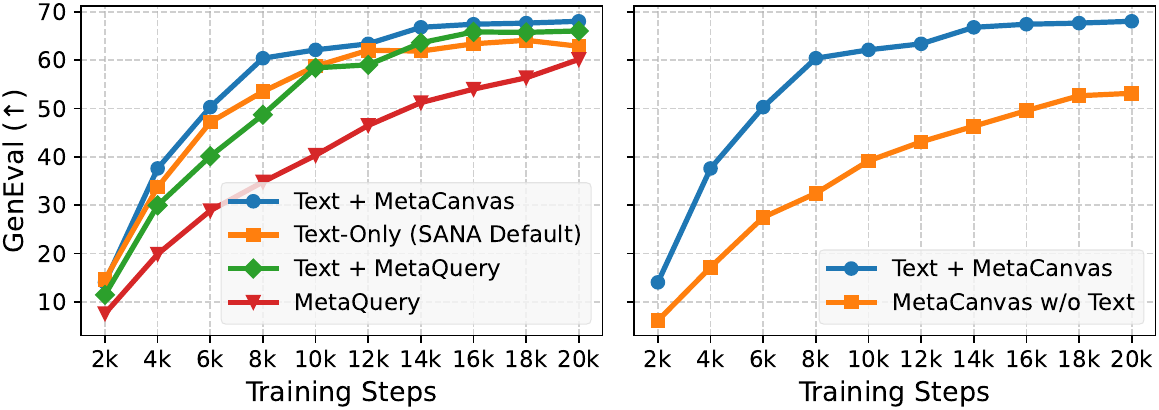}
  \vspace{-1mm}
  \caption{
  \textbf{Left:} Comparison of \method with MetaQuery~\citep{pan2025metaqueries} and text conditioning. \textbf{Right:} Comparison of \method with and without additional text conditioning.
  \vspace{-0mm}
  }
\label{fig:geneval_curves} 
\end{figure}

\begin{figure*}[t]
  \centering
  \includegraphics[width=0.95\linewidth]{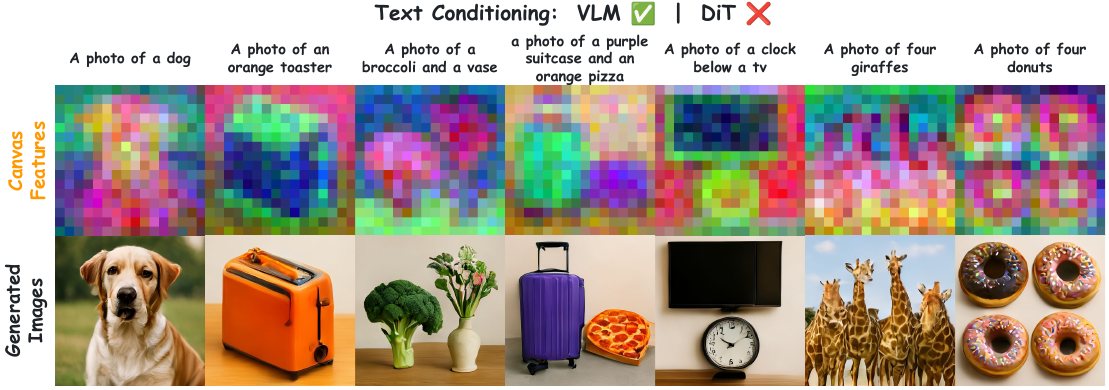}
  \vspace{-1mm}
  \caption{
\textbf{Visualization of canvas features (1st row) and generated images (2nd row) {\color{purple}using only canvas tokens without extra text conditioning in DiT.}}
We train SANA~\citep{xie2024sana} {\sl from scratch} using only canvas tokens from Qwen2.5-VL~\citep{Qwen2-VL} as the conditioning input, {\sl with no text signals provided to the DiT}. Following~\citep{tumanyan2023plug}, we apply PCA to the features produced by the \method{} connector.
Canvas tokens output from MLLM can serve as reasonable visual planning sketches to effectively guide the final image synthesis in the DiT.
  }
  \vspace{-0mm}
\label{fig:attention_map_visualization} 
\end{figure*}

\begin{table*}[t]
\begin{center}
\setlength{\tabcolsep}{0.32em}
\renewcommand{\arraystretch}{1.0}
\caption{{\textbf{Quantitative comparison with models on ImgEdit~\citep{ye2025imgedit}  benchmark.} 
Full table is shown in \Cref{table:results_imgedit}.}}
\label{table:results_imgedit_short}
\vspace{-1mm}
\scalebox{0.8}{
\begin{tabular}{l|cccccccccc}
\hlineB{3}
{Method} & {Add} & {Adjust} & {Extract} & {Replace} & {Remove} & {Background} & {Style} & {Hybrid} & {Action} & {Overall$\uparrow$} \\  
\hline
 Step1X-Edit~\citep{liu2025step1x} & 3.88 & 3.14 & 1.76 & 3.40 & 2.41 & 3.16 & 4.63 & 2.64 & 2.52 & 3.06 \\
 BAGEL~\citep{deng2025emerging} & 3.56 & 3.31 & 1.70 & 3.30 & 2.62 & 3.24 & 4.49 & 2.38 & 4.17 & 3.20 \\
 OmniGen2~\citep{wu2025omnigen2} & 3.57 & 3.06 & 1.77 & 3.74 & 3.20 & 3.57 & 4.81 & 2.52 & 4.68 & 3.44 \\
 GPT-Image-1 [High]~\citep{openai_image_api_2025} & {4.61} & 4.33 & 2.90 & 4.35 & 3.66 & 4.57 & 4.93 & 3.96 & 4.89 & 4.20 \\
 Qwen-Image-Edit-2509~\citep{wu2025qwen} & 4.32 & 4.36 & 4.04 & 4.64 & 4.52 & 4.37 & 4.84 & 3.39 & 4.71 & 4.35 \\
\hline
 FLUX.1 Kontext [Dev]~\citep{batifol2025flux} & 3.76 & 3.45 & 2.15 & 3.98 & 2.94 & 3.78 & 4.38 & 2.96 & 4.26 & 3.52 \\
 \rowcolor{lightblue}
{FLUX.1 Kontext [Dev] + MetaCanvas}  & 4.20 & 3.50 & 2.11 & 4.41 & 3.72 & 3.89 & 4.83 & 3.61 & 4.49 & 3.86      \\ 
 \rowcolor{lightblue}
& \textcolor{blue}{0.44$\uparrow$} & \textcolor{blue}{0.05$\uparrow$}& \textcolor{red}{0.04$\downarrow$}& \textcolor{blue}{0.43$\uparrow$}& \textcolor{blue}{0.78$\uparrow$}& \textcolor{blue}{0.11$\uparrow$}& \textcolor{blue}{0.45$\uparrow$}& \textcolor{blue}{0.65$\uparrow$}& \textcolor{blue}{0.23$\uparrow$}& \textcolor{blue}{0.34$\uparrow$}\\
\hlineB{3}
\end{tabular}}
\end{center}
\end{table*}

\begin{table}[t]
\begin{center}
\setlength{\tabcolsep}{0.49em}
\renewcommand{\arraystretch}{1.0}
\caption{{\textbf{Quantitative comparison results on GEdit-EN-full~\citep{liu2025step1x} benchmark.} Best numbers are \textbf{bolded}, and the second best are \underline{underlined}. Full table is shown in \Cref{table:results_gedit}.
}}
\label{table:results_gedit_short}
\vspace{-1mm}
\scalebox{0.9}{
\begin{tabular}{l|ccc}
\hlineB{3}
\multirow{2}{*}{\textbf{Model}} & \multicolumn{3}{c}{\textbf{GEdit-Bench-EN$\uparrow$}} \\ \cline{2-4}  
 & \textbf{G\_SC} & \textbf{G\_PQ} & \textbf{G\_O} \\  
\hline
 UniWorld-V1~\citep{lin2025uniworld} & 4.93 & 7.43 & 4.85 \\
 Step1X-Edit (v1.1)~\citep{liu2025step1x} & 7.66 & 7.35 & 6.97 \\
 BAGEL~\citep{deng2025emerging} & 7.36 & 6.83 & 6.52 \\
 OmniGen2~\citep{wu2025omnigen2} & 7.16 & 6.77 & 6.41 \\
  GPT-Image-Edit~\citep{wang2025gpt} & 8.00 & \textbf{7.86} & 7.56 \\
  Qwen-Image-Edit-2509~\citep{wu2025qwen} & \textbf{8.80} & \underline{7.74} & \textbf{7.98} \\
\hline
 FLUX.1-Kontext-Dev~\citep{batifol2025flux} & 6.52 & 7.38 & 6.00 \\
 \rowcolor{lightblue}
FLUX.1-Kontext-Dev + MetaCanvas  &  \underline{8.24} & 7.68 & \underline{7.67}   \\ 
 \rowcolor{lightblue}
& \textcolor{blue}{1.72$\uparrow$} & \textcolor{blue}{0.50$\uparrow$} & \textcolor{blue}{1.67$\uparrow$} \\
\hlineB{3}
\end{tabular}}
\end{center}
\end{table}

\begin{figure}[t]
    \vspace{-0mm}
    \centering
    \includegraphics[width=0.7\linewidth]{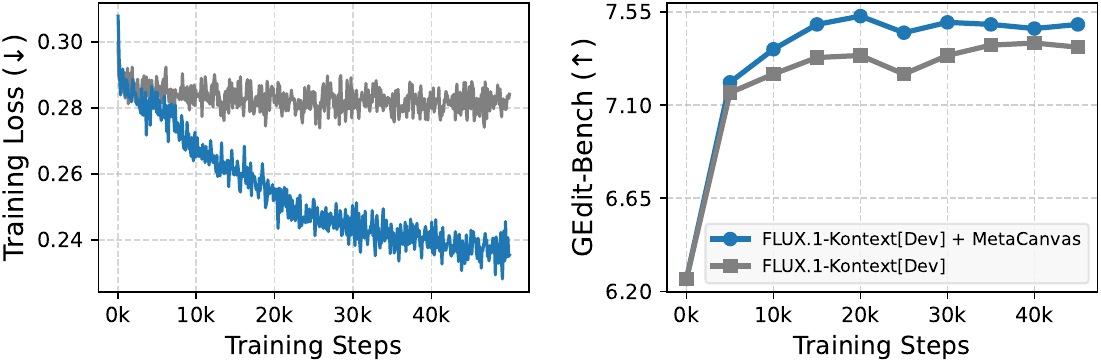}
    \vspace{-1mm}
    \caption{
        Comparison of training loss and GEdit-Bench~\citep{liu2025step1x} scores for the baseline method without canvas tokens and \method. Both models are fine-tuned on the same training dataset.
    }
    \label{fig:gedit_curve}
    \vspace{-0mm}
\end{figure}

\section{Results and Discussion}
\label{sec:results}

We empirically validate \method across multiple tasks and diffusion model backbones. Specifically, we begin with exploratory T2I generation experiments as proof of concept (\Cref{subsec:results_exploratory_experiments_t2i}), then implement the connector design on image editing task (\Cref{subsec:results_image_editing}), and finally scale up to video generation and editing tasks (\Cref{subsec:results_video_generation_editing}).

\subsection{Exploratory Experiments on T2I Generation}
\label{subsec:results_exploratory_experiments_t2i}

We aim to validate two questions here: {\color{orange}\textbf{Q1:}} Does \method{} really help guide the generation process of diffusion models? {\color{orange}\textbf{Q2:}} What connector design is most effective?

\thinparagraph{Comparison with other design choices.} To answer {\color{orange}\textbf{Q1}}, in \Cref{fig:geneval_curves} (left), we compare \method with (1) the default SANA baseline (T5 text conditioning), (2) an architecture equivalent to MetaQuery~\citep{pan2025metaqueries} that uses 256 learnable 1D query tokens produced by Qwen-2.5-VL while reusing the same text-conditioning interface, and (3) a variant that concatenates T5 text embeddings with the 256 MetaQuery tokens for additional context. As shown, combining text as global guidance with \method as a visual prior yields consistent gains and has the fastest GenEval convergence among all variants.
In \Cref{fig:geneval_curves} (right), we further evaluate a no-text variant. Even without any text conditioning, adding 2D learnable canvas tokens on top of the noisy latents in DiT provides meaningful structural guidance, demonstrating \textit{effective information transfer from the MLLM to the DiT via \method}. Visualization of this no-text variant on GenEval examples is shown in~\Cref{fig:attention_map_visualization}.

\begin{wraptable}[11]{r}{0.5\linewidth}
\begin{center}
\setlength{\tabcolsep}{0.49em}
\renewcommand{\arraystretch}{1.0}
\caption{{\textbf{Ablation study on \method connector design.}}}
\label{table:geneval_method_ablations}
\vspace{-2mm}
\scalebox{0.85}{
\begin{tabular}{l|c}
\hlineB{3}
{} & {\textbf{GenEval$\uparrow$}}  \\ 
\hline
 \textbf{\method} (ours) & 68.02 \hspace{1mm}\color{white}{(0.00$\downarrow$)} \\
 \hspace{3mm}Remove Timestep Condition in DiT Block & 67.42 \hspace{1mm}\color{Red}{(0.60$\downarrow$)} \\
 \hspace{3mm}Remove DiT Block & 66.39 \hspace{1mm}\color{Red}{(1.03$\downarrow$)} \\
 \hspace{3mm}Remove Vanilla Transformer Block & 66.19 \hspace{1mm}\color{Red}{(0.20$\downarrow$)} \\
 \hspace{3mm}Add Canvas Tokens Before Patchification & 65.34 \hspace{1mm}\color{Red}{(0.85$\downarrow$)} \\ 
 \hline
 \textbf{Baseline} (Default SANA Architecture) & 64.09 \hspace{1mm}\color{white}{(0.00$\downarrow$)} \\ 
\hlineB{3}
\end{tabular}}
\end{center}
\end{wraptable}
\thinparagraph{Connector design.} We address {\color{orange}\textbf{Q2}} with an ablation study on the \method connector design in~\Cref{table:geneval_method_ablations}. Visualizations of these architectural variants are provided in the appendix. We find that conditioning on the timestep enables dynamic control over the influence of canvas tokens on the noisy latents, while the proposed DiT block and accompanying transformer blocks effectively transform and fuse canvas-token information with the latents. Moreover, avoiding early projection of canvas tokens into the low-dimensional VAE space yields additional gains.

\subsection{Results on Image Editing Task}
\label{subsec:results_image_editing}

We evaluate the fine-tuned image-editing model FLUX.1-Kontext [Dev]~\citep{batifol2025flux} augmented with \method against competing methods on ImgEdit-Bench (see \Cref{table:results_imgedit_short}) and GEdit-Bench (see \Cref{table:results_gedit_short}). Equipping FLUX.1-Kontext [Dev] with \method yields consistent improvements on both benchmarks. 
\Cref{fig:gedit_curve} further contrasts the vanilla model with its \method-augmented counterpart under the same training setup, showing steady gains throughout training. Notably, these benefits come from adding only lightweight connector modules, incurring minimal parameter and computational overhead.

\subsection{Results on Video Generation and Editing Tasks}
\label{subsec:results_video_generation_editing}

\begin{table}[h]
\begin{center}
\setlength{\tabcolsep}{0.49em}
\renewcommand{\arraystretch}{1.0}
\caption{{\textbf{Quantitative comparison results on VBench-I2V~\citep{huang2024vbench++}.} Best numbers are \textbf{bolded}, and the second best are \underline{underlined}. Full table is shown in the Appendix.
}}
\label{table:results_t2v_i2v}
\vspace{-1mm}
\scalebox{0.8}{
\begin{tabular}{lccc}
\hlineB{3}
 & \textbf{I2V Score$\uparrow$} & \textbf{Quality Score$\uparrow$} & \textbf{Overall$\uparrow$} \\  
\hline
 CogVideoX-5B~\citep{kong2024hunyuanvideo} & 94.79 & \underline{78.61} & 86.70  \\
 HunyuanVideo~\citep{kong2024hunyuanvideo} & 95.10 & 78.54 & 86.82  \\
 Wan2.1-14B~\citep{wan2025wan} & 92.90 & \textbf{80.82} & 86.86  \\
 Wan2.2-5B~\citep{wan2025wan} & \underline{95.69} & 78.26  & \underline{86.98}  \\
\hline
 \rowcolor{lightblue}
Wan2.2-5B + MetaCanvas & \textbf{97.50} & 76.76  & \textbf{87.13}   \\ 
\hlineB{3}
\end{tabular}}
\end{center}
\end{table}

\thinparagraph{Video generation.} In~\Cref{table:results_t2v_i2v}, we compare videos generated by \method{} after three training stages (see~\Cref{subsec:exp_setup_training}) with open-source models, including CogVideoX-5B~\citep{yang2024cogvideox}, HunyuanVideo~\citep{kong2024hunyuanvideo}, Wan2.1-14B~\citep{wan2025wan}, and the baseline model Wan2.2-5B~\citep{wan2025wan}.
Our method achieves comparable performance while being \emph{additionally equipped} with strong video editing capabilities.

\begin{table*}[h]
\begin{center}
\setlength{\tabcolsep}{0.65em}
\renewcommand{\arraystretch}{1.0}
\caption{{{\textbf{Quantitative comparison on video editing task.} The best and second best numbers are \textbf{bolded} and \underline{underlined}.}}}
\label{table:results_video_editing}
\vspace{-1mm}
\scalebox{0.76}{
\begin{tabular}{llccccccc}
\hlineB{3}
\multirow{2}{*}{\textbf{Model}} & \multirow{2}{*}{\textbf{DiT Backbone}} & {\textbf{VBench}} & \multicolumn{3}{c}{\textbf{VLM Eval (GPT-4o)}} & \multicolumn{3}{c}{\textbf{Human Preference Rate}}  \\  \cline{4-6} \cline{3-3} \cline{7-9}  
 & & \textbf{Quality$\uparrow$} & \textbf{Sementics$\uparrow$} & \textbf{Quality$\uparrow$} & \textbf{Overall$\uparrow$} & \textbf{Edit Acc.$\uparrow$} & \textbf{Consistency$\uparrow$} & \textbf{Overall$\uparrow$} \\  
\hline
 \multicolumn{2}{l}{\textcolor{gray}{Methods w/ Text Encoder}} \\
 InsViE~\citep{wu2025insvie} & CogVideoX-2B & 79.19 & 3.85 & 4.48 & 3.60 & - & - & - \\
 Lucy-Edit-Dev~\citep{decart2025lucyedit} & Wan2.2-5B & 78.07 &  5.57 & 7.19 & 5.44 & - & - & - \\
 Lucy-Edit-Dev v1.1~\citep{decart2025lucyedit} & Wan2.2-5B & 77.70 & \underline{5.68} & 6.59 & 5.43 & 9.5\% & \underline{42.7\%} & \underline{26.10\%} \\
 Ditto~\citep{bai2025scaling} & Wan2.1-14B & \underline{80.64} &  4.90 & \textbf{7.54} & {5.48} & \underline{18.4\%} & 7.7\% & 13.04\% \\
\hline
 \multicolumn{2}{l}{\textcolor{gray}{Method w/ MLLM as Multimodal Encoder}} \\
 w/o Canvas Tokens (Baseline) & Wan2.2-5B & \textbf{81.39} & \underline{6.61} & {7.25} & \underline{6.68} & {-} & -  & -  \\ 
 \rowcolor{lightblue}
w/ Canvas Tokens (\method{}) & Wan2.2-5B & {80.27} & \textbf{7.91} & \underline{7.50} & \textbf{7.56} & \textbf{72.1\%} & \textbf{49.6\%}  & \textbf{60.84\%}  \\ 
\hlineB{3}
\end{tabular}}
\end{center}
\end{table*}

\begin{figure*}[t]
  \centering
  \includegraphics[width=0.99\linewidth]{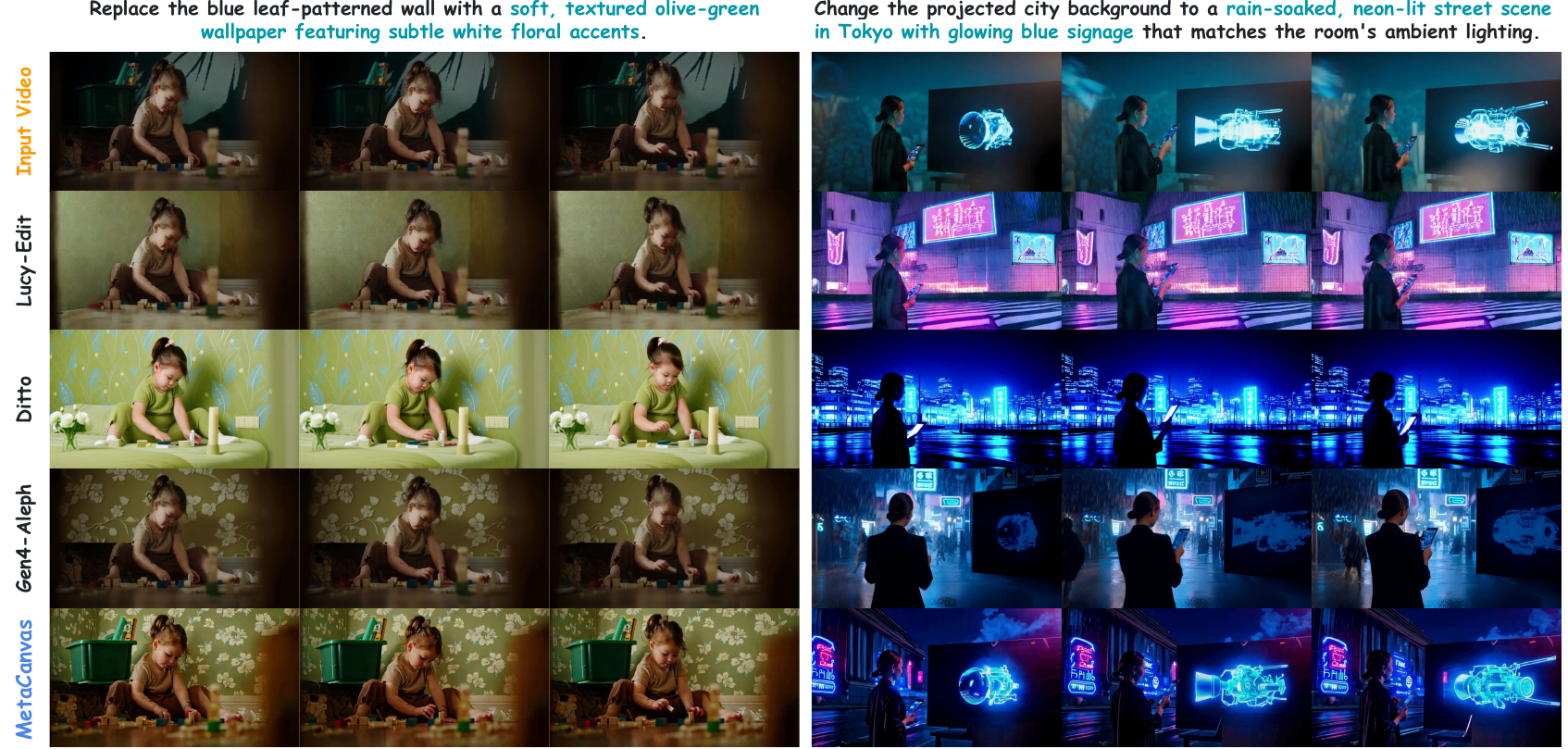} 
  \vspace{-1mm}
  \caption{
  \textbf{Visualization on video editing task}. \method{} demonstrates strong prompt understanding and effective grounding of the editing regions, achieving better spatial alignment and improved preservation of both foreground and background details.
  }
\label{fig:visualization_video_editing} 
\end{figure*}

\thinparagraph{Video editing.} We compare \method{} with recent state-of-the-art models, including InsViE~\citep{wu2025insvie}, Ditto~\citep{bai2025scaling}, and Lucy-Edit-Dev~\citep{decart2025lucyedit}, as well as a control setup of our method that excludes canvas tokens. As shown in~\Cref{table:results_video_editing}, \method{} achieves comparable video quality scores, as measured by VBench~\citep{huang2023vbench, huang2024vbench++} and GPT-4o~\citep{openai2024gpt4o}, while outperforming all baselines in editing accuracy (i.e., semantics) by a large margin. In addition, we conduct human evaluations comparing Lucy-Edit-DeV v1.1, Ditto, and \method{}, and report the win rates for editing accuracy, spatio-temporal consistency, and overall user preference (see Appendix for details). \method{} achieves the highest preference rate across all evaluation dimensions. Furthermore, the controlled variant without canvas tokens attains competitive or better performance relative to other baselines, demonstrating the effectiveness of replacing the text encoder with a MLLM-based multimodal condition encoder. We provide visualization of these methods together with closed-source model Gen4-Aleph~\citep{runwayml_gen4_aleph_2025} in~\Cref{fig:visualization_video_editing}. \method{} demonstrates strong prompt understanding and effective grounding of the editing regions, with better spatial alignment and preservation of foreground and background details.

\begin{table}[h!]
\begin{center}
\setlength{\tabcolsep}{0.65em}
\renewcommand{\arraystretch}{1.0}
\caption{{Comparison between models with and without Canvas tokens for video editing task, and ablations of \method with different 
number of canvas keyframes. Best and second best numbers are \textbf{bolded} and \underline{underlined}.}
} 
\label{table:video_editing_keyframes_ablations}
\vspace{-1mm}
\scalebox{0.9}{
\begin{tabular}{lcccccc}
\hlineB{3}
\multirow{2}{*}{} & \textbf{w/o Canvas} & \multicolumn{5}{c}{\textbf{Canvas w/ Different \# Keyframes}} \\
\cline{3-7}
&  \textbf{Tokens} & {\cellcolor{lightblue}1} & {\cellcolor{lightblue}3} & 6 & 11 & 31 \\
\hline
\multicolumn{2}{l}{\textcolor{gray}{Spatial Alignment}} & {\cellcolor{lightblue}} & {\cellcolor{lightblue}} \\
Flow Err.$\downarrow$ & 6.07 & {\cellcolor{lightblue}\textbf{4.92}} & {\cellcolor{lightblue}\underline{4.91}} & 5.48  & 5.25 & 5.59 \\
PSNR$\uparrow$ & 12.36 & {\cellcolor{lightblue}\underline{13.62}} & {\cellcolor{lightblue}\textbf{13.65}} & 13.04 & 13.08 & 12.54 \\
\hline
\multicolumn{2}{l}{\textcolor{gray}{Video Quality Evaluation}}  & {\cellcolor{lightblue}} & {\cellcolor{lightblue}} \\
VBench$\uparrow$ & 
\textbf{81.39} & 
{\cellcolor{lightblue}79.75} & {\cellcolor{lightblue}80.28} & 79.92 & {79.91} & \underline{80.79} \\
\hline
\multicolumn{2}{l}{\textcolor{gray}{VLM Eval (GPT-4o)}}  & {\cellcolor{lightblue}} & {\cellcolor{lightblue}} \\
Sementics$\uparrow$ & 6.61 & {\cellcolor{lightblue}\textbf{8.01}} & {\cellcolor{lightblue}\underline{7.91}} & 7.46 & 7.60 & 7.35 \\  
Quality$\uparrow$ & 7.25 & {\cellcolor{lightblue}\textbf{7.85}} & {\cellcolor{lightblue}\underline{7.50}} & 7.25 & 7.45 & 7.26\\  
Overall$\uparrow$ & 6.68 & {\cellcolor{lightblue}\textbf{7.82}} & {\cellcolor{lightblue}\underline{7.56}} & 7.20 & 7.36 & 7.12 \\  
\hline
\multicolumn{2}{l}{\textcolor{gray}{Train Time / Step (Seconds)}}  & {\cellcolor{lightblue}} & {\cellcolor{lightblue}} \\
 & 20.75 & {\cellcolor{lightblue}21.40} & {\cellcolor{lightblue}21.46} & 21.77 & 22.08 & 23.12 \\  
\hlineB{3}
\end{tabular}}
\end{center}
\end{table}

\thinparagraph{Ablation study on the number of canvas keyframes.}
Our default design uses three canvas keyframes for video tasks. As an ablation, we compare different numbers of canvas keyframes in~\Cref{table:video_editing_keyframes_ablations}. We find that using three keyframes achieves better performance than using more keyframes, highlighting the simplicity and efficiency of our design. Although using a single keyframe (i.e., 2D canvas) slightly outperforms three keyframes in MLLM-based evaluation, we empirically observe noticeable temporal flickering in the first few frames, resulting in a lower VBench video quality score (79.75 vs. 80.28). We hypothesize that this is due to the VAE design in Wan2.2-5B, which we describe in more detail in~\Cref{subsec:appendix_2d_vs_3d_canvas}. Compared with the controlled variant without canvas tokens, adding three canvas keyframes increases the clock time per training step by only 3.1\% (from 20.75 to 21.40 seconds per step).

\thinparagraph{In-context video generation.}
In~\Cref{table:results_omnicontext_video}, we compare \method{} with Wan-VACE~\citep{jiang2025vace}-1.3B/14B on video generation from reference images. \method{} achieves competitive performance with these baselines, particularly on human-object interaction tasks (i.e., character + object under multiple ID categories). Visualization results for \method{} and Wan-VACE-14B are shown in~\Cref{fig:ri2v_visualization} and in the Appendix. We describe our training data curation pipeline in the Appendix, and acknowledge that there remains room for further improvement in our data creation strategy.

\begin{figure}[t]
  \centering
  \includegraphics[width=0.55\linewidth]{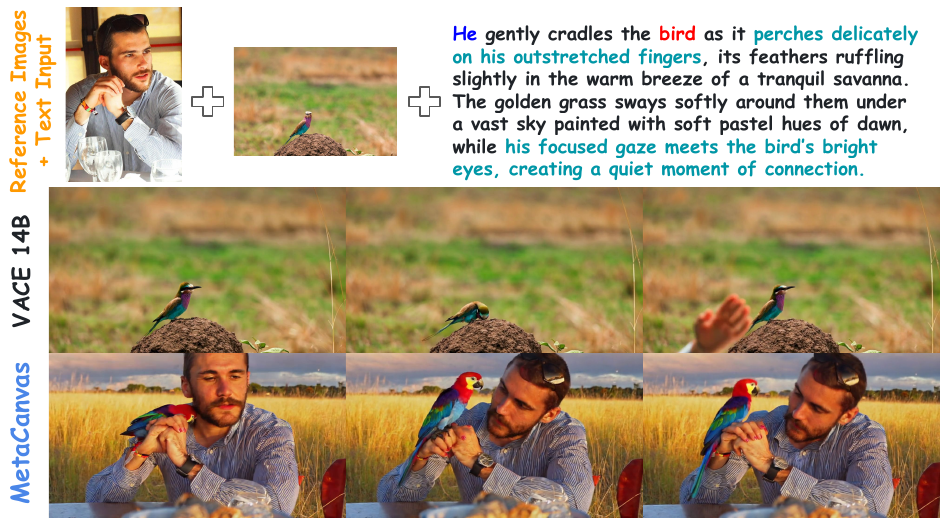}
  \vspace{-1mm}
  \caption{
  \textbf{Visualization of in-context video generation.} \method{} accurately generates the video by composing reference images in accordance with the text prompt.
  }
\label{fig:ri2v_visualization} 
\end{figure}

\begin{table*}[t!]
\begin{center}
\setlength{\tabcolsep}{0.7em}
\renewcommand{\arraystretch}{1.0}
\caption{{\textbf{Quantitative comparison on OmniContext-Video benchmark for in-context video generation from reference images.}}}
\label{table:results_omnicontext_video}
\vspace{-1mm}
\scalebox{0.78}{
\begin{tabular}{l|cc|ccc|cccc}
\hlineB{3}
\multirow{2}{*}{\textbf{Model}} & \multicolumn{2}{c}{\textbf{Single ID}} & \multicolumn{3}{c}{\textbf{Multiple IDs}} & \multicolumn{3}{c}{\textbf{Scene}} & \multirow{2}{*}{\textbf{Average$\uparrow$}}  \\ \cline{2-9} 
 & \textbf{Char.} & \textbf{Obj.} & \textbf{Char.} & \textbf{Obj.} & \textbf{Char. + Obj.} & \textbf{Char.} & \textbf{Obj.} & \textbf{Char. + Obj.} \\  
\hline
 Wan-VACE-1.3B~\citep{jiang2025vace} & 6.44  & 4.88 & \textbf{4.36} & 5.66 & 4.36 & 3.37 & 4.88 & \textbf{4.92} & 4.85 \\
 Wan-VACE-14B~\citep{jiang2025vace} & \textbf{6.71} & 5.79 & 3.16 & 5.10 & 4.64 & 3.85 & 4.00 & 4.64 & 4.86 \\
\hline
 \rowcolor{lightblue}
Wan2.2-5B~\citep{wan2025wan} + MetaCanvas  & 6.16 & \textbf{6.34} & 3.77 & \textbf{5.69} & \textbf{6.35} & \textbf{4.60} & \textbf{5.43} & 4.75 & \textbf{5.40}     \\ 
\hlineB{3}
\end{tabular}}
\end{center}
\end{table*}

\section{Conclusion}
\label{sec:conclusion}
In this paper, we present \method, an elegant framework that bridges MLLMs and diffusion models through a novel connector design. By introducing learnable multi-dimensional canvas tokens as spatiotemporal priors that fuses multimodal input conditions, \method{} effectively plans and guides media synthesis in a principled manner. Across three backbones and six diverse tasks (image/video generation, editing, and in-context generation with multimodal cues), our experiments show that \method{} delivers strong performance while maintaining training efficiency relative to prior architectures.

\clearpage
\newpage
\bibliographystyle{assets/plainnat}
\bibliography{paper}

\clearpage
\newpage
\beginappendix

\appendix

\crefalias{section}{appendix}
\crefalias{subsection}{appendix}

\startcontents[sections]
\printcontents[sections]{l}{1}{\setcounter{tocdepth}{3}}

\section{Background}

\subsection{Extended Related Works}
\label{sec:appendix_related_works}

\thinparagraph{Native models for visual generation and understanding.}
There has been extensive efforts in recent models to unify visual and text generation and understanding capabilities into a single model. The visual generation capabilites in these models are trained with either the same autoregressive objectives as text, such as Chameleon~\citep{ChameleonEarlyFusion2024}, VILA-U~\citep{wu2024vilau}, TokenFlow~\citep{geyertokenflow}, EMU3~\citep{wang2024emu3}, JanusPro~\citep{chen2025januspro}, or through diffusion denoising or flow-matching objective, such as Transfusion~\citep{zhou2024transfusion}, BAGEL~\citep{deng2025emerging}, JanusFlow~\citep{ma2024janusflow}, Show-o~\citep{xie2024showo}, LlamaFusion~\citep{shi2025lmfusion}, Mogao~\citep{liao2025mogao}.
{On the positive side, these approaches are designed to maximize modality and task fusion through end-to-end (E2E) training, which usually initialize the backbone from a (M)LLM, or a dual-branch architecture that models text and vision. With potential higher upper bound due to E2E model designs, such methods require significant resources to learn high-quality visual generation from scratch.} {Unlike native approaches, MetaCanvas bridges the merits of existing MLLMs and diffusion generators without extensive retraining.}

\thinparagraph{Bridging (M)LLMs with diffusion models.}
Another approach is to bridge a pretrained (M)LLM with a strong pretrained diffusion model, preserving the strengths of {the individual components}. The simplest variant is text only guidance: the LLM expands or structures the prompt (e.g., detailed captions, layouts), which then conditions the diffusion model~\citep{li2023gligen, yang2023reco, feng2023layoutgpt, Cho2023VPT2I, linvideodirectorgpt, zaladiagrammergpt, lianllm}. 
{While efficient, pure text is often too sparse to convey the dense visual cues required for complex scenes with high fidelity}.
A more expressive alternative is to directly use the output token embeddings from the MLLM~\citep{pan2025metaqueries, tong2024metamorph, liu2025step1x, lin2025bifrost, wu2025omnigen2, lin2025uniworld, koh2023GILL, ge2024seedx, chen2025blip3ofamilyfullyopen, pankosmos, yin2025best, song2025query, mou2025instructx}. In this setting, the MLLM provides the final hidden states of the multimodal inputs, or a sequence of learned query embeddings {summarizing the multimodal conditioning}; these are injected into the diffusion model {via various conditioning injection methods, including the self or cross attention (aka text conditioning method), token concatenation, or lightweight adapters like control nets}. Representative examples include GILL~\citep{koh2023GILL}, which maps LLM features into the diffusion model text encoder space, and MetaMorph~\citep{tong2024metamorph}, which trains an MLLM to autoregressively produce image condition tokens later consumed by diffusion. MetaQueries~\citep{pan2025metaqueries} and BLIP3-o~\citep{chen2025blip3ofamilyfullyopen} learn a fixed set of query vectors and a transformer based connector that {abstract} MLLM semantics to condition the generator.
Although this paradigm is generally more compute efficient than training a native unified architecture and tends to preserve the MLLM’s reasoning ability, encoding structural visual priors into a single-dimensional (1D) token sequence is non-trivial. Models often require large connector {blocks} and substantial data to recover precise positional {and temporal} information. Consequently, 1D tokens can be insufficient to model composition, object placement, geometry {and motion}, motivating interfaces with explicit {multi-dimensional} grounding.
Bifrost-1~\citep{lin2025bifrost} addresses this challenge by leveraging patch-level latents that are natively aligned with the MLLM visual encoder as 2D visual priors (rather than compressing the information into a 1D token sequence) to guide the diffusion model. This design outperforms architectures that rely on non-MLLM-aligned visual features (e.g., VAE, SigLIP) and significantly improves training efficiency. However, fine-tuning the visual generation branch, implemented as a trainable copy initialized from the original MLLM parameters (e.g., Qwen2.5-VL-7B), becomes increasingly challenging as the MLLM’s parameter size grows, indicating that there remains room for architectural improvement.

\thinparagraph{Summary.} In contrast to {previous work, we propose a novel design of} learnable multidimensional canvas tokens that are trained jointly with a small connector and injected directly into the diffusion latent {space}, creating a compact and extensible framework that {excels in explicitly composing spatial temporal signatures for diffusion generation. Our method is capable to generalize from images (2D) to videos (3D) seamlessly, while also maintaining efficiency in the volume of canvas tokens in the interface}. 
{In summary, \method preserves the strengths of each backbone, namely understanding for the MLLMs and generation for diffusion models, lowers the training cost, and scales naturally across image and video tasks, effectively serving as a SOTA-level lightweight bridge from MLLMs to diffusion models.}

\begin{figure*}[t!]
  \centering
  \includegraphics[width=0.95\linewidth]{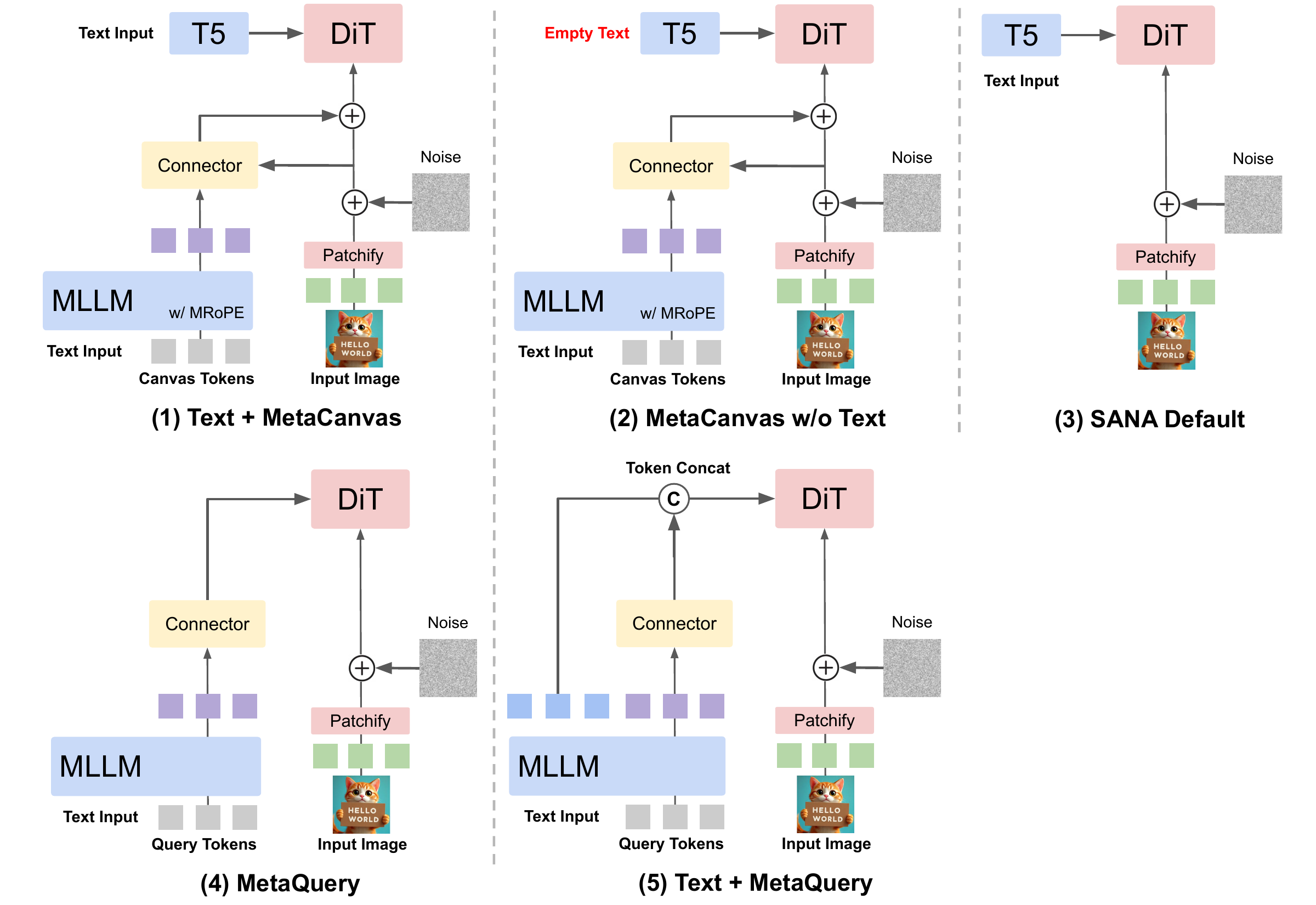}
  \caption{
  \textbf{Visualizations of the architectures for the comparison variants in~\Cref{fig:geneval_curves}.}
  }
\label{fig:appendix_comparisoin_with_other_architectures} 
\end{figure*}

\begin{figure*}[t!]
  \centering
  \includegraphics[width=0.95\linewidth]{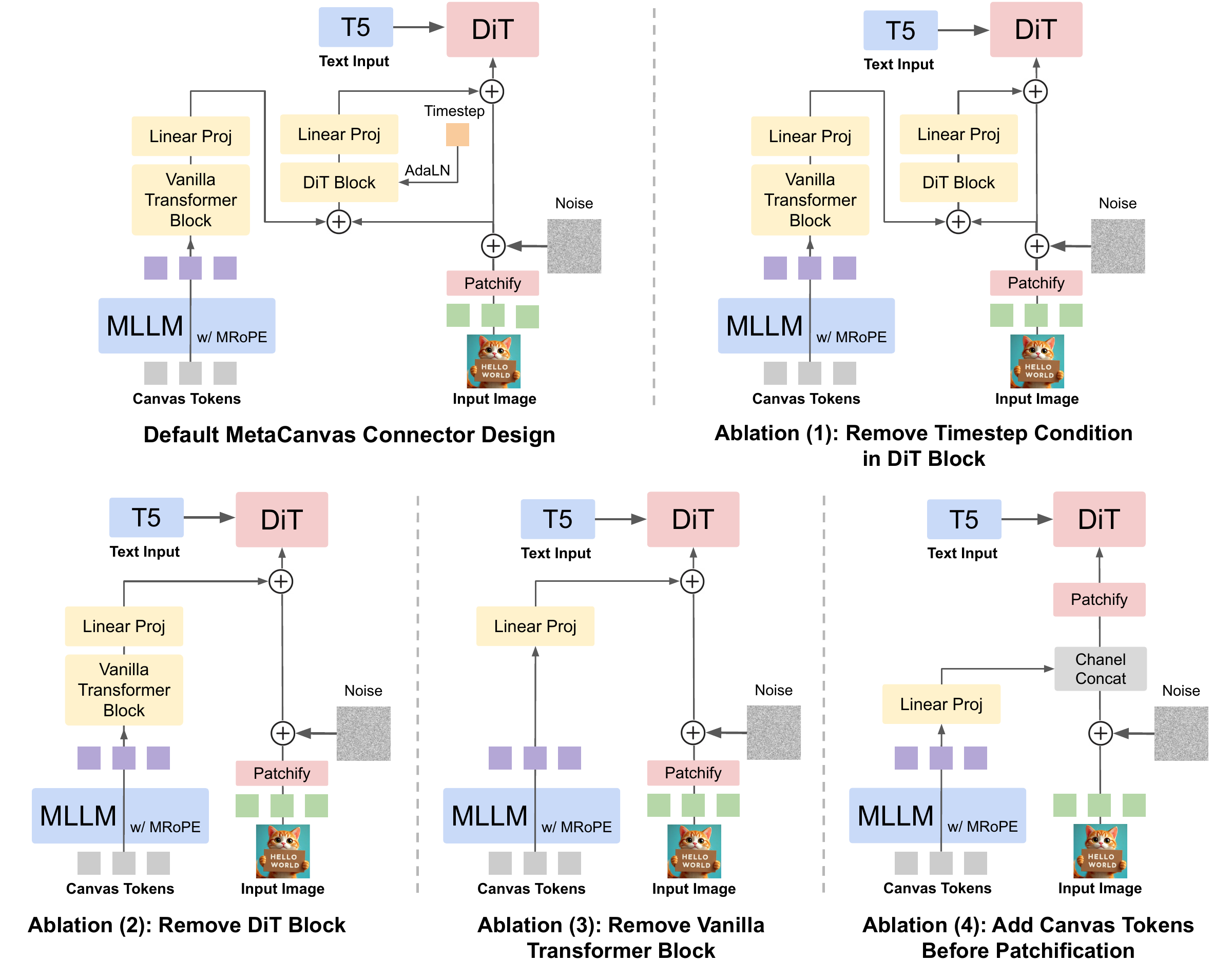}
  \vspace{-1mm}
  \caption{
  \textbf{Ablation study on \method connector architecture design. Corresponding quantitative results are shown in \Cref{fig:connector_design}.} Note that our focus here is to demonstrate the effectiveness of canvas tokens; therefore, we keep the T5 text encoder in SANA instead of replacing it with text embeddings from MLLMs. We later replace the text encoder with the output multimodal embeddings from the MLLM in the image editing and video tasks in~\Cref{subsec:results_image_editing} and~\Cref{subsec:results_video_generation_editing}.
  }
\label{fig:appendx_canvas_connector_ablation} 
\end{figure*}

\subsection{Extended Preliminaries}
\label{sec:appendix_preliminaries}

\textbf{Latent Diffusion Models with Flow-Matching.}
Given an RGB visual input $\bm{x}\in\mathbb{R}^{F\times 3\times H\times W}$, with $F=1$ for images and $F>1$ for videos, we first encode it with a spatial (or spatio\-temporal) variational autoencoder (VAE)~\citep{kingma2013auto} into $d$-dimensional latents $\bm{z}=\bm{\mathcal{E}}(\bm{x})\in\mathbb{R}^{F'\times d\times H'\times W'}$, where $F'\leq F$, $H'<H$, and $W'<W$ due to downsampling~\citep{Qwen2.5-VL, internvl3, meta2025llama4, google2025gemini25pro}.
We then adopt a flow-matching objective~\citep{lipmanflow}.
Specifically, let $\bm{\epsilon}\sim\mathcal{N}(\bm{0},\bm{I})$ and $t\sim\mathcal{U}(0,1)$. 
We define a straight-line path (rectified flow) in latent space
$
\bm{z}_t=(1-t)\,\bm{z}+t\,\bm{\epsilon},
$
and the corresponding target velocity field
$
\bm{u}_t=\frac{d\bm{z}_t}{dt}=\bm{\epsilon}-\bm{z}.
$
A latent diffusion model (LDM) $\bm{\mathcal{F}}_\theta(\bm{z}_t, t, \bm{c}_{\text{text}}, \bm{c}_{\text{vision}})$ is trained to match this field given conditioning text $\bm{c}_{\text{text}}$ and/or images/videos $\bm{c}_{\text{vision}}$, via
$
\mathcal{L}_{\text{FM}}
=\mathbb{E}_{\bm{z},\bm{\epsilon},t}\left\|
\bm{\mathcal{F}}_\theta(\bm{z}_t, t, \bm{c}_{\text{text}}, \bm{c}_{\text{vision}})
-\big(\bm{\epsilon}-\bm{z}\big)
\right\|_2^2$.
At inference, we solve the ODE $\,\tfrac{d\bm{z}_t}{dt}=\bm{\mathcal{F}}_\theta(\bm{z}_t, t, \bm{c}_{\text{text}}, \bm{c}_{\text{vision}})\,$ from $t=1$ to $t=0$ with multiple denoising steps to obtain $\hat{\bm{z}}$, and decode with the VAE decoder $\hat{\bm{x}}=\bm{\mathcal{D}}(\hat{\bm{z}})$. 
In \method, we train with the same flow matching objective $\mathcal{L}_{\text{FM}}$ without extra auxiliary losses, and we use the same denoising strategy during inference.

\section{Exploratory T2I Generation}
\label{sec:appendix_exploratory_t2i_generation}

\subsection{Architecture Details of Other Design Choices}

Here, we provide a more detailed illustration of the experiments in~\Cref{fig:geneval_curves} of~\Cref{subsec:results_exploratory_experiments_t2i}.
Specifically, in this exploratory experiment, we aim to demonstrate that:
\begin{itemize}[leftmargin=1em]
    \setlength{\itemsep}{0pt}
    \setlength{\parskip}{0pt}
    \setlength{\parsep}{0pt}
    \item Canvas tokens can effectively transfer information from the MLLM into DiT.
    \item The transferred information provides additional gains beyond the information contained in text.
    \item Combining canvas information with text leads to faster training convergence compared with other information injection strategies, including MetaQuery~\citep{pan2025metaqueries}.
\end{itemize}

The quantitative results for these variants are shown in~\Cref{fig:geneval_curves}, and their architectures are visualized in~\Cref{fig:appendix_comparisoin_with_other_architectures}:
\begin{itemize}[leftmargin=1em]
    \setlength{\itemsep}{0pt}
    \setlength{\parskip}{0pt}
    \setlength{\parsep}{0pt}
    \item \textbf{(1) Text + MetaCanvas:} We use the canvas embeddings output from the MLLM, together with the text embeddings from the original T5 text encoder in SANA as conditioning. Our goal is not to create a unified conditioning strategy by replacing the original T5 text encoder with an MLLM; rather, we only add the canvas tokens output from the MLLM to the latents before inputting them into DiT.
    \item \textbf{(2) MetaCanvas w/o Text:} We provide empty text (i.e., Text Input = ``) to the T5 text encoder. This allows us to control for the same DiT architecture without removing any components from its cross-attention blocks.
    \item \textbf{(3) SANA Default:} This is the default architecture used in SANA, where only text information is used for conditional generation.
    \item \textbf{(4) MetaQuery:} The 1D query embeddings output from the MLLM are provided as input, using the same cross-attention interface of DiT.
    \item \textbf{(5) Text + MetaQuery:} We concatenate the text embeddings from the T5 text encoder with the learnable query tokens from the MLLM and input them into DiT to provide richer information.
\end{itemize}

\subsection{Ablation for Canvas Connector Design}

In~\Cref{fig:appendx_canvas_connector_ablation}, we provide a more detailed visualization of the ablation study for canvas connector architecture design variants, as mentioned in~\Cref{table:geneval_method_ablations} of~\Cref{subsec:method_connector_design}. The effectiveness of each component is quantitatively demonstrated in~\Cref{table:geneval_method_ablations}, so we skip redundant illustration here for simplicity.

\subsection{Experiment Setup Details}

In~\Cref{table:t2i_training_hyperparameters}, we give the hyper-parameter and model details of the experiments in~\Cref{fig:geneval_curves} and~\Cref{table:geneval_method_ablations}. 
Specifically, in both experiments, we keep the parameters of the MLLM fully frozen without adding any LoRA, and we train the DiT \emph{from scratch}.

Additionally, as the generated images are of $512\times512$ resolution and the compression rate is 32, the DiT latents have a shape of $16\times16$. Therefore, we use canvas tokens of the same shape, resulting in a total of 256 canvas tokens. The training is conducted on a single A100 node and takes around one day to complete for approximately 20k steps.

\begin{figure*}[t!]
  \centering
  \includegraphics[width=0.8\linewidth]{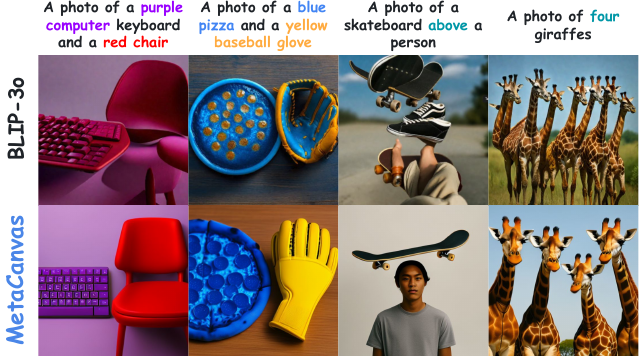}
  \caption{
  \textbf{Comparison between \method{} with query-based architecture (i.e., BLIP-3o~\citep{chen2025blip3ofamilyfullyopen}) on GenEval~\citep{ghosh2023geneval} prompts.} 
  }
\label{fig:appendix_comparison_blip3o} 
\end{figure*}

\subsection{Additional Comparison with Query-Based Architecture}

In our main paper, we compare \method{} with MetaQuery by training the DiT (i.e., SANA) \emph{from scratch}. Here, we provide additional experiments that compare \method{} with this line of query-based architectures. Since MetaQuery is not open-sourced, we compare our method with BLIP-3o~\citep{chen2025blip3ofamilyfullyopen}, another model that uses query tokens generated from an MLLM as conditioning for the DiT. For a fair comparison, our model is trained solely on the BLIP3o-60k dataset, which is a subset of the training data used in BLIP-3o. In this experiment, we apply LoRA fine-tuning to the DiT with a rank of 32.

In~\Cref{fig:appendix_comparison_blip3o}, we compare \method{} with BLIP-3o on several challenging GenEval prompts, including attribute combinations with uncommon colors (e.g., ``A photo of a blue pizza and a yellow baseball glove''), unusual spatial relationships (e.g., ``A photo of a skateboard above a person''), and object counts greater than three (e.g., ``A photo of four giraffes''). As shown in the figure, \method{} consistently generates images that correctly capture these attributes, spatial relationships, and counts, whereas BLIP-3o fails on these prompts by mixing colors between objects, struggling to produce counterfactual spatial relationships such as a person below a skateboard, and generating an incorrect number of giraffes.

In~\Cref{table:comparison_geneval_sota}, we compare our model with models that use a DiT-only architecture for generation-only tasks, as well as models with a query-based architecture, including MetaQuery and BLIP-3o. Note that BLIP3o-8B achieves better performance than MetaQuery-XL, while our model uses the same MLLM and DiT (i.e., Qwen2.5-VL-7B and SANA1.0-1.6B) as MetaQuery, and is trained on BLIP3o-60k, which is a subset of the training data used in BLIP-3o. Under this controlled setting, our method still outperforms both query-based models. These quantitative results, together with the visualizations above, demonstrate the effectiveness of our canvas design.

\begin{table}[t!]
\begin{center}
\setlength{\tabcolsep}{0.99em}
\renewcommand{\arraystretch}{1.0}
\caption{\small{\textbf{Quantitative results on GenEval benchmarks.} Here, we LoRA finetune SANA instead of training it from scratch. For a fair comparison, \method{} is trained solely on the BLIP3o-60k dataset, which is a subset of the training data used in BLIP-3o.}}
\label{table:comparison_geneval_sota}
\vspace{-2mm}
\scalebox{0.99}{
\begin{tabular}{lc}
\hlineB{3}
\textbf{Model} & \textbf{GenEval} \\
\hline
\multicolumn{2}{l}{\textcolor{gray}{DiT w/o MLLM}}
\\
FLUX1-Dev~\citep{flux} & 0.82  \\
SANA~\citep{xie2024sana} & 0.66 \\
\hline
\multicolumn{2}{l}{\textcolor{gray}{Query-Based Architecture}}
\\
MetaQuery-XL~\citep{pan2025metaqueries} & 0.80    \\
BLIP3o-8B~\citep{chen2025blip3ofamilyfullyopen} & 0.84   \\
\hline
\rowcolor{lightblue}
\method{} (Ours) & \textbf{0.86}    \\
\hlineB{3}
\end{tabular}}
\end{center}
\end{table}

\begin{table}[h!]
\begin{center}
\setlength{\tabcolsep}{0.7em}
\renewcommand{\arraystretch}{1.0}
\caption{\small{\textbf{Hyper-parameter and training details of the experiments in~\Cref{fig:geneval_curves} and~\Cref{table:geneval_method_ablations}.}}}
\label{table:t2i_training_hyperparameters}
\vspace{-1mm}
\scalebox{0.95}{
\begin{tabular}{lc}
\hlineB{3}
Learning Rate & 1e-4  \\
Warmup Learning Rate & 1e-5 \\
LR Scheduler & Constant w/ WarmUp    \\
Weight Decay & 0.0   \\
Gradient Clip & 1.0    \\
Optimizer & AdamW  \\
Warm-Up Steps & 300    \\
Training Steps & 20k   \\
Image Batch Size / GPU & 40   \\
\# A100 GPUs & 8  \\
MLLM Training Type & Frozen \\
DiT Training Type & From Scratch \\
\# Canvas Tokens & 16$\times$16$=$256  \\
\hlineB{3}
\end{tabular}}
\end{center}
\end{table}

\begin{table}[h!]
\begin{center}
\setlength{\tabcolsep}{0.9em}
\renewcommand{\arraystretch}{1.0}
\caption{\small{\textbf{Ablation study on the effectiveness of multimodal-RoPE for encoding learnable canvas tokens.} 
}}
\label{table:appendix_ablation_mrope}
\vspace{-2mm}
\scalebox{0.99}{
\begin{tabular}{lc}
\hlineB{3}
 & \textbf{GenEval} \\
\hline
\method{} w/o MRoPE & {0.85}    \\
\rowcolor{lightblue}
\method{} (Default) & \textbf{0.86}    \\
\hlineB{3}
\end{tabular}}
\vspace{-2mm}
\end{center}
\end{table}

\subsection{Additional Ablation Results}
\label{subsec:appendix_additional_ablation_results}

\thinparagraph{Multimodal RoPE in MLLM.} Here, we discuss the use of multimodal RoPE in the MLLM to process the learnable canvas tokens. Specifically, for image and video understanding tasks in Qwen2.5-VL~\citep{Qwen2.5-VL}, images and videos are encoded using multimodal RoPE (i.e., MRoPE), which provides improved spatial–temporal encoding of visual information. Therefore, we adopt the same strategy to encode our learnable canvas tokens. In~\Cref{table:appendix_ablation_mrope}, we show that \method{} without multimodal RoPE achieves slightly worse results on GenEval compared with our default setting, although both variants outperform previous models, including MetaQuery and BLIP-3o. Since applying multimodal RoPE is essentially free that requires only a preprocessing step for position IDs and incurring no additional computational cost, we adopt it as our default configuration.

\noindent\textbf{MLLM LoRA.} In addition, we observe that adding a LoRA with rank of 32 to the MLLM effectively increases its capacity and helps the canvas tokens extract information more accurately from the MLLM, which further improves the GenEval score from 0.86 to 0.87.

\section{Image Editing Task}
\label{sec:appendix_image_editing_task}

\begin{figure}[t]
  \centering
  \includegraphics[width=0.55\linewidth]{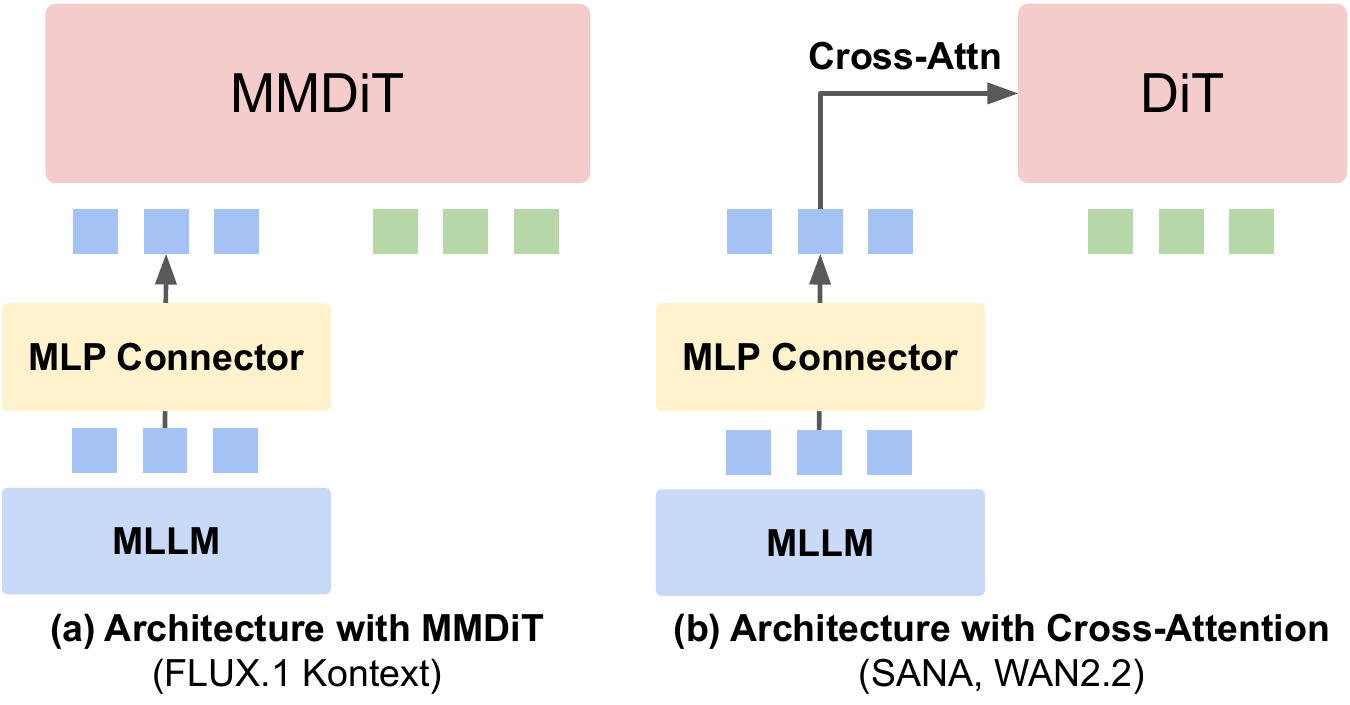}
  \caption{
  \textbf{\method implemented with MMDiT and cross-attention-based architectures.}
  }
\label{fig:mmdit_crossattn} 
\end{figure}

\begin{figure*}[t]
  \centering
  \includegraphics[width=0.99\linewidth]{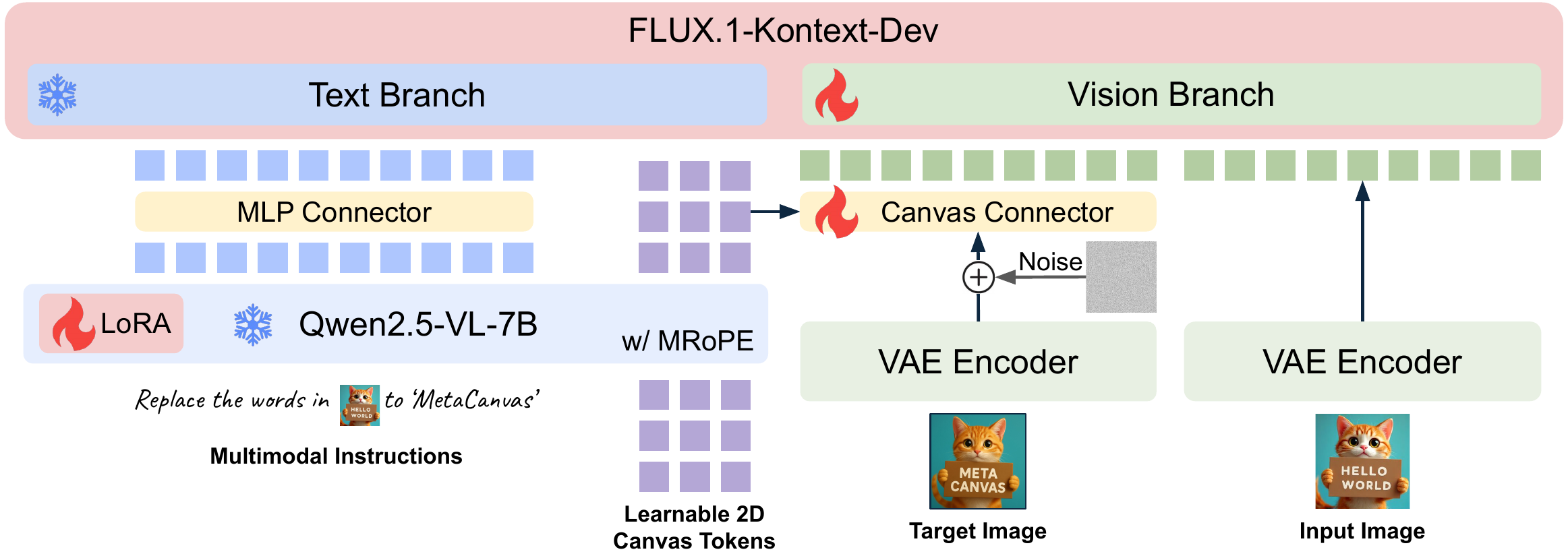}
  \caption{
  \textbf{\method implementation architecture details in~\Cref{subsec:results_image_editing}.} We adopt FLUX.1-Kontext-Dev~\citep{batifol2025flux} as the MMDiT and Qwen2.5-VL-7B~\citep{Qwen2.5-VL} as the MLLM. The trainable components include the vision branch, the LoRA in the MLLM, as well as the canvas tokens and the corresponding lightweight canvas connector.
  }
\label{fig:appendix_image_editing_architecture} 
\end{figure*}

\begin{table*}[h!]
\begin{center}
\setlength{\tabcolsep}{0.4em}
\renewcommand{\arraystretch}{1.0}
\caption{{\textbf{Quantitative comparison results on GEdit-EN-full~\citep{liu2025step1x} benchmark.} 
}}
\label{table:results_gedit}
\vspace{-1mm}
\scalebox{0.76}{
\begin{tabular}{l|cccccccccccc}
\hlineB{3}
{Method} & {BG} & {Color} & {Mat.} & {Motion} & {Portrait} & {Style} & {Add} & {Remove} & {Replace} & {Text} & {Tone} & {Avg$\uparrow$} \\  
\hline
 Instruct-Pix2Pix~\citep{brooks2023instructpix2pix} & 3.94 & 5.40 & 3.52 & 1.27 & 2.62 & 4.39 & 3.07 & 1.50 & 3.48 & 1.13 & 5.10 & 3.22 \\
 MagicBrush~\citep{zhang2023magicbrush} & 6.17 & 5.41 & 4.75 & 1.55 & 2.90 & 4.10 & 5.53 & 4.13 & 5.10 & 1.33 & 5.07 & 4.19 \\
 OmniGen~\citep{xiao2025omnigen} & 5.23 & 5.93 & 5.44 & 3.12 & 3.17 & 4.88 & 6.33 & 6.35 & 5.34 & 4.31 & 4.96 & 5.01 \\
 Step1X-Edit~\citep{liu2025step1x} & 7.03 & 6.26 & 6.46 & 3.66 & 5.23 & 7.24 & 7.17 & 6.42 & 7.39 & 7.40 & 6.62 & 6.44 \\
 Step1X-Edit (v1.1)~\citep{liu2025step1x} & 7.45 & 7.38 & 6.95 & 4.73 & 4.70 & 7.11 & 8.20 & 7.59 & 7.80 & 7.91 & 6.85 & 6.97 \\
 BAGEL~\citep{deng2025emerging} & 7.44 & 6.99 & 6.26 & 5.09 & 4.82 & 6.04 & 7.94 & 7.37 & 7.31 & 7.16 & 6.17 & 6.60 \\
 OmniGen2~\citep{wu2025omnigen2} & - & - & - & - & - & - & - & - & - & - & - & 6.42 \\
  GPT-Image-Edit~\citep{wang2025gpt} & 7.80 & 7.54 & 7.12 & 7.75 & 7.09 & 6.74 & 8.04 & 7.95 & 7.17 & 5.45 & 6.95 & 7.24 \\
 Qwen-Image-Edit-2509~\citep{wu2025qwen} & 8.25 & 8.10 & 7.51 & 8.60 & 6.92 & 6.85 & 8.56 & 8.59 & 8.37 & 8.18 & 7.85 & 7.98 \\
\hline
 FLUX.1-Kontext-Dev~\citep{batifol2025flux} & 7.06 & 7.03 & 5.52 & 5.62 & 4.68 & 5.55 & 6.95 & 6.76 & 6.13 & 6.10 & 7.48 & 6.26 \\
 \rowcolor{lightblue}
FLUX.1-Kontext-Dev + MetaCanvas  & 7.71 & 7.88 & 7.27 & 7.97 & 7.82 & 6.93 & 8.44 & 8.15 & 8.02 & 6.11 & 8.06 & 7.67     \\ 
 \rowcolor{lightblue}
 & \textcolor{blue}{0.65$\uparrow$} & \textcolor{blue}{0.85$\uparrow$}& \textcolor{blue}{1.75$\uparrow$}& \textcolor{blue}{2.35$\uparrow$}& \textcolor{blue}{3.14$\uparrow$}& \textcolor{blue}{1.38$\uparrow$}& \textcolor{blue}{1.49$\uparrow$}& \textcolor{blue}{1.39$\uparrow$}& \textcolor{blue}{1.89$\uparrow$}& \textcolor{blue}{0.01$\uparrow$}& \textcolor{blue}{0.58$\uparrow$}& \textcolor{blue}{1.41$\uparrow$}\\
\hlineB{3}
\end{tabular}}
\end{center}
\end{table*}

\begin{table*}[h!]
\begin{center}
\setlength{\tabcolsep}{0.3em}
\renewcommand{\arraystretch}{1.0}
\caption{{\textbf{Quantitative comparison results on ImgEdit~\citep{ye2025imgedit}  benchmark.} 
}}
\label{table:results_imgedit}
\vspace{-1mm}
\scalebox{0.8}{
\begin{tabular}{l|cccccccccc}
\hlineB{3}
{Method} & {Add} & {Adjust} & {Extract} & {Replace} & {Remove} & {Background} & {Style} & {Hybrid} & {Action} & {Overall$\uparrow$} \\  
\hline
 MagicBrush~\citep{zhang2023magicbrush} & 2.84 & 1.58 & 1.51 & 1.97 & 1.58 & 1.75 & 2.38 & 1.62 & 1.22 & 1.90 \\
 Instruct-Pix2Pix~\citep{brooks2023instructpix2pix} & 2.45 & 1.83 & 1.44 & 2.01 & 1.50 & 1.44 & 3.55 & 1.20 & 1.46 & 1.88 \\
 UltraEdit~\citep{zhao2024ultraedit} & 3.44 & 2.81 & 2.13 & 2.96 & 1.45 & 2.83 & 3.76 & 1.91 & 2.98 & 2.70 \\
 OmniGen~\citep{xiao2025omnigen} & 3.47 & 3.04 & 1.71 & 2.94 & 2.43 & 3.21 & 4.19 & 2.24 & 3.38 & 2.96 \\
 ICEdit~\citep{zhang2025context} & 3.58 & 3.39 & 1.73 & 3.15 & 2.93 & 3.08 & 3.84 & 2.04 & 3.68 & 3.05 \\
 Step1X-Edit~\citep{liu2025step1x} & 3.88 & 3.14 & 1.76 & 3.40 & 2.41 & 3.16 & 4.63 & 2.64 & 2.52 & 3.06 \\
 BAGEL~\citep{deng2025emerging} & 3.56 & 3.31 & 1.70 & 3.30 & 2.62 & 3.24 & 4.49 & 2.38 & 4.17 & 3.20 \\
 UniWorld-V1~\citep{lin2025uniworld} & 3.82 & 3.64 & 2.27 & 3.47 & 3.24 & 2.99 & 4.21 & 2.96 & 2.74 & 3.26 \\
 OmniGen2~\citep{wu2025omnigen2} & 3.57 & 3.06 & 1.77 & 3.74 & 3.20 & 3.57 & 4.81 & 2.52 & 4.68 & 3.44 \\
 GPT-Image-Edit~\citep{wang2025gpt} & 4.07 & 3.79 & 2.04 & 4.13 & 3.89 & 3.90 & 4.84 & 3.04 & 4.52 & 3.80 \\
 FLUX.1 Kontext [Pro]~\citep{batifol2025flux} & 4.25 & 4.15 & 2.35 & 4.56 & 3.57 & 4.26 & 4.57 & 3.68 & 4.63 & 4.00 \\
 GPT-Image-1 [High]~\citep{openai_image_api_2025} & 4.61 & 4.33 & 2.90 & 4.35 & 3.66 & 4.57 & 4.93 & 3.96 & 4.89 & 4.20 \\
 Qwen-Image-Edit~\citep{wu2025qwen} & 4.38 & 4.16 & 3.43 & 4.66 & 4.14 & 4.38 & 4.81 & 3.82 & 4.69 & 4.27 \\
 Qwen-Image-Edit-2509~\citep{wu2025qwen} & 4.32 & 4.36 & 4.04 & 4.64 & 4.52 & 4.37 & 4.84 & 3.39 & 4.71 & 4.35 \\
\hline
 FLUX.1 Kontext [Dev]~\citep{batifol2025flux} & 3.76 & 3.45 & 2.15 & 3.98 & 2.94 & 3.78 & 4.38 & 2.96 & 4.26 & 3.52 \\
 \rowcolor{lightblue}
FLUX.1 Kontext [Dev] + MetaCanvas  & 4.20 & 3.50 & 2.11 & 4.41 & 3.72 & 3.89 & 4.83 & 3.61 & 4.49 & 3.86      \\ 
 \rowcolor{lightblue}
 & \textcolor{blue}{0.44$\uparrow$} & \textcolor{blue}{0.05$\uparrow$}& \textcolor{red}{0.04$\downarrow$}& \textcolor{blue}{0.43$\uparrow$}& \textcolor{blue}{0.78$\uparrow$}& \textcolor{blue}{0.11$\uparrow$}& \textcolor{blue}{0.45$\uparrow$}& \textcolor{blue}{0.65$\uparrow$}& \textcolor{blue}{0.23$\uparrow$}& \textcolor{blue}{0.34$\uparrow$}\\
\hlineB{3}
\end{tabular}}
\end{center}
\end{table*}

As FLUX.1-Dev~\citep{flux} and FLUX.1-Kontext-Dev~\citep{batifol2025flux} share the same T5 text encoder, we follow GPT-Image-Edit~\citep{wang2025gpt} and initialize from the MLP connector in UniWorld-V1~\citep{lin2025uniworld}, which was trained to bridge Qwen2.5-VL-7B~\citep{Qwen2.5-VL} and FLUX.1-Dev~\citep{flux}. Note that SANA uses cross-attention as its conditioning interface, whereas FLUX.1-Dev concatenates text and image tokens as input to the DiT. This architectural difference leads to differences in how text tokens are injected when replacing the original text encoder of the DiT with output embeddings from the MLLM. A high-level illustration comparing the MMDiT and cross-attention–based architectures is provided in~\Cref{fig:mmdit_crossattn}. A more detailed version that specifically bridges FLUX.1-Kontext-Dev with Qwen2.5-VL-7B is shown in~\Cref{fig:appendix_image_editing_architecture}.

We start the training directly by unfreezing the diffusion model’s visual branch, together with the additional learnable canvas tokens and the transformer-based connector. To increase MLLM's model capacity, we apply LoRA with rank of 64 to the MLLM backbone. We illustrate the training hyper-parameters in~\Cref{table:appendix_image_editing_hyperparameters}.
The model is trained on $\mathcal{O}$(1M) image-editing samples from OmniEdit~\citep{wei2024omniedit} and HQEdit~\citep{hui2024hq}. \Cref{table:results_imgedit} and~\Cref{table:results_gedit} contain extended evaluation results of~\Cref{table:results_imgedit_short} and~\Cref{table:results_gedit_short}.

\begin{table}[h!]
\begin{center}
\setlength{\tabcolsep}{0.8em}
\renewcommand{\arraystretch}{1.0}
\caption{\small{\textbf{Hyperparameters and training details of the image editing experiments in~\Cref{table:results_imgedit_short} and~\Cref{table:results_gedit_short}.}}}
\label{table:appendix_image_editing_hyperparameters}
\vspace{-1mm}
\scalebox{0.85}{
\begin{tabular}{lc}
\hlineB{3}
\multicolumn{2}{l}{\textcolor{gray}{Training Hyperparameters}}
\\
Learning Rate & 1e-6  \\
LR Scheduler & Constant    \\
Adam $\beta_1$ & 0.9 \\
Adam $\beta_2$ & 0.99 \\
Adam $\epsilon$ & 1e-8 \\
Weight Decay & 0.0   \\
Gradient Clip & 1.0    \\
Optimizer & AdamW  \\
Warm-Up Steps & 0    \\
Training Steps & 90k   \\
Image Batch Size / GPU & 1   \\
\# A100 GPUs & 64  \\
\hline
\multicolumn{2}{l}{\textcolor{gray}{Model Parameters}}
\\
MLLM Training Type & LoRA  \\
MLLM LoRA Rank & 64 \\
MLLM LoRA Dropout & 0.05 \\
DiT Training Type & Train Vision Branch \\
\# Canvas Tokens & 32$\times$32$=$1024  \\
\hlineB{3}
\end{tabular}}
\end{center}
\end{table}

\section{\method{} Architecture Design for Video Tasks}

\subsection{Context and Canvas Connectors Design}
\label{subsec:appendix_context_canvas_connectors_design}

\thinparagraph{Context connector design.} For the MLP context connector in video tasks, we adopt a similar strategy as in the image editing tasks by training a 2-layer MLP, with input channels equal to the embedding dimension of the MLLM and output channels equal to the embedding dimension of the DiT. We use an expansion ratio of 4 in the middle layer of the MLP to increase its expressiveness. A dropout ratio of 0.1 is applied to the parameters in this MLP connector.

\thinparagraph{Canvas connector design.} 
For the canvas connector, we use the same architecture as in the image generation and editing tasks without any modifications. Specifically, the canvas connector consists of a single vanilla transformer block and a single DiT block. The vanilla transformer block is designed to more effectively transform the canvas embeddings output by the MLLM into the feature space of the DiT, while the DiT block further fuses the canvas embeddings with the DiT latents and dynamically determines the influence of the canvas tokens on the latents through AdaLN. As the self-attention module in the DiT operates spatially and temporally across all frames, the canvas connector here can be applied only on a per-frame basis rather than across all frames, which reduces GPU memory consumption.

\subsection{Interface Design for Canvas Keyframe and Reference/Condition Frames Injection}

One key design aspect is the interface for injecting canvas keyframes into the DiT while seamlessly enabling diverse video tasks, including text-to-video generation, image-to-video generation, video editing, and reference-to-video generation. Our goal is to design an interface that is compatible with all of these tasks and can be easily extended to additional video tasks.
We partially follow the interface of Wan-Animate~\citep{cheng2025wan} to support conditioning on both reference frames and source video frames. In addition, to effectively inject canvas embeddings into the latents, we use the design shown in~\Cref{fig:video_patch_embed_design} to combine the canvas embeddings with the latents after patchification.

\begin{figure}[t]
  \centering
  \includegraphics[width=0.65\linewidth]{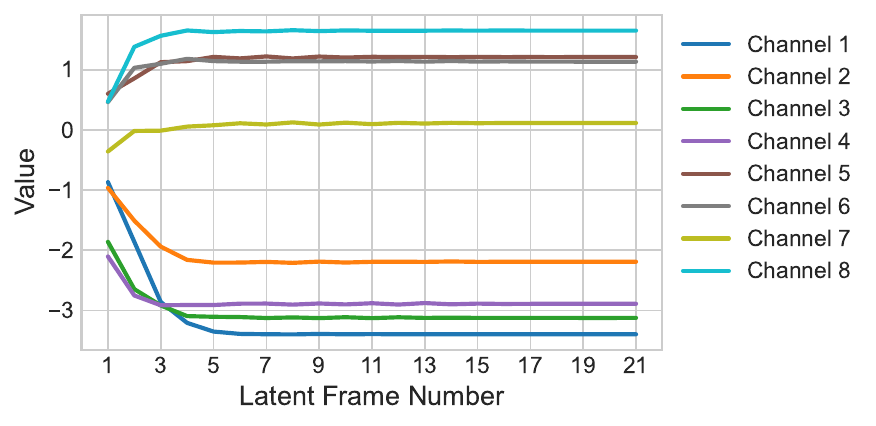}
  \vspace{-1mm}
  \caption{
  \textbf{Visualization of the first 8 channels of Wan2.2 VAE after encoding a \emph{static} video of 81 frames (21 latent frames).}
  }
\label{fig:visualization_vae} 
\end{figure}

\begin{table*}[t!]
\begin{center}
\setlength{\tabcolsep}{0.4em}
\renewcommand{\arraystretch}{1.0}
\caption{{\textbf{Quantitative comparison results on VBench-I2V~\citep{huang2024vbench++} for I2V generation.} 
}
}
\label{table:results_vbench_i2v}
\vspace{-1mm}
\scalebox{0.82}{
\begin{tabular}{l|cccccccccc}
\hlineB{3}
\multirow{2}{*}{\textbf{Method}} & \textbf{I2V} & \textbf{I2V} & \textbf{Camera} & \textbf{Subject} & \textbf{Background} & \textbf{Motion} & \textbf{Dynamic} & \textbf{Aesthetic} & \textbf{Imaging} & \multirow{2}{*}{\textbf{Overall$\uparrow$}} \\  
& \textbf{Subject} & \textbf{BG.} & \textbf{Motion} & \textbf{Consistency} & \textbf{Consistency} & \textbf{Smooth.} & \textbf{Degree} & \textbf{Quality} & \textbf{Quality} & \\  
\hline
 Wan-5B & 97.79 & 98.92 & 42.54 & 94.31 & 96.14 & 96.60 & 58.53 & 61.95 & 71.24 & 86.98 \\
 \rowcolor{lightblue}
 Wan-5B + \method & 97.89 & 98.81 & 80.60 & 94.24 & 93.66 & 97.32 & 54.30 & 60.97 & 70.38 & 87.13 \\
\hlineB{3}
\end{tabular}}
\end{center}
\end{table*}

\subsection{2D canvas \emph{vs.} 3D canvas}
\label{subsec:appendix_2d_vs_3d_canvas}

\thinparagraph{Quantitative comparison.} A natural question is whether we should keep the same 2D canvas for video tasks, as in image tasks, or instead use a 3D canvas to encode both spatial and temporal information. As shown quantitatively in~\Cref{table:video_editing_keyframes_ablations}, using a 2D canvas achieves comparable spatial alignment scores (measured by optical flow error and PSNR) and also attains the best evaluation score with GPT-4o. Although the overall evaluation scores remain reasonably high, we empirically observe that using a 2D canvas introduces noticeable temporal flickering in the first few frames, whereas applying a 3D canvas with three keyframes effectively alleviates this issue.

\thinparagraph{Hypothesis and explanations.} 
Upon further investigation, we found that this discrepancy likely stems from the VAE used in Wan2.2. Specifically, the Wan2.2 VAE supports both image and video encoding: the first frame is encoded as a single latent frame, while every subsequent four frames are collapsed into one latent frame. Additionally, the VAE employs a moving-average mechanism to better preserve temporal information across latent frames.
To illustrate this behavior, we encode a \textbf{static} 81-frame video using the Wan2.2 VAE, resulting in 21 latent frames. \Cref{fig:visualization_vae} visualizes the first eight channels of these latents. We observe clear value inconsistencies across the initial latent frames, while the later latent frames become stable. This explains why a 2D canvas, whose embeddings are added identically to all latent frames, produces temporal flickering in the early frames. In contrast, a 3D canvas with 3 keyframes adapts to the temporal structure of the latent sequence and effectively mitigates this issue. We acknowledge that the current design may still be suboptimal and leave the exploration of improved designs for future work.

\section{Text/Image-to-Video Generation Tasks}

\textbf{Task-specific model details.}
For the image-to-video generation task, we follow the setting in Wan2.2-5B by replacing the first latent frame in the noisy latents with the clean latent of the first input image. During both training and inference, the first latent frame is assigned a timestep of zero, while the remaining latent frames receive noise according to their corresponding timesteps. This ensures that the generated video better preserves the information from the input image.

\noindent\textbf{Training dataset details.} In stage 1 (a), we use approximately $\mathcal{O}(40\text{M})$ in-house images for training. In later stages, we further incorporate approximately $\mathcal{O}(8\text{M})$ in-house videos for training.

\section{Video Editing Task}
\label{sec:appendix_video_editing_task}

\subsection{Training dataset details.} 

As there's no natual source-edited video pairs, we utilize the following strategy to construct a total of $300k$ video editing samples. Specifically, our training data comes from two sources:
\begin{itemize}[leftmargin=1em]
    \setlength{\itemsep}{0pt}
    \setlength{\parskip}{0pt}
    \setlength{\parsep}{0pt}
    \item \textbf{Generate from videos.} Given a raw video, we use a MLLM (i.e., Qwen2.5-VL) to generate a video editing instruction that performs one of the following operations: local object addition, replacement, deletion, background change, or global style transfer. We then use Qwen-Image-Edit~\citep{wu2025qwen} to generate the edited first frame. Additionally, we employ depth ControlNet to extract depth maps from the raw video. Finally, the edited video is generated from the edited first frame and the corresponding depth maps.  
    \item \textbf{Generate from images.} Starting from paired image editing data, we use Wan2.2-5B~\citep{wan2025wan} to animate the raw image into a video and extract depth frames via depth ControlNet. The edited video is then generated from the edited first frame and the extracted depth maps.
\end{itemize}

\subsection{Evaluation benchmark details.} 

\noindent\textbf{Evaluation dataset curation strategy.} Due to the scarcity of open-source video editing evaluation benchmarks with high resolution (at least 720p) and sufficient duration (at least 121 frames at 24 FPS), we curate a balanced evaluation set consisting of 300 video prompts. This set covers a diverse range of video editing tasks, including local object editing (addition, removal, or replacement), background change, and style transfer. Specifically, we first randomly sample 3k videos from our high-quality in-house video data, which are not used during training to avoid potential data leakage. For each video, we prompt GPT-4o~\citep{openai_gpt4o_2024} to generate an editing instruction corresponding to one of the target categories, ensuring that the instructions balance creativity with natural suitability to the video content. Next, GPT-4o is used to filter out low-quality video-editing prompt pairs and to maintain a balanced distribution of prompts across categories. We also perform a manual verification step to ensure the instructions are meaningful and suitable to the videos. Additionally, half of the videos contain humans, which helps align the evaluation set with user needs for human-centered video generation.

\begin{table*}[t!]
\begin{center}
\setlength{\tabcolsep}{0.95em}
\renewcommand{\arraystretch}{1.0}
\caption{Quantitative comparison results on our OmniContext-Video benchmark for in-context video generation from reference images. This table is an extended version of~\Cref{table:results_omnicontext_video} that further includes detailed prompt-following and subject-consistency scores for completeness, in addition to the overall score. Compared with \method{}, VACE-14B~\citep{jiang2025vace} tends to over-copy the reference images into the final generated video, making it harder to follow the user instructions.}
\label{table:appendix_results_omnicontext_video}
\vspace{-1mm}
\scalebox{0.78}{
\begin{tabular}{l|cc|ccc|cccc}
\hlineB{3}
\multirow{2}{*}{\textbf{Prompt-Following}} & \multicolumn{2}{c}{{Single ID}} & \multicolumn{3}{c}{{Multiple IDs}} & \multicolumn{3}{c}{{Scene}} & \multirow{2}{*}{{Average$\uparrow$}}  \\ \cline{2-9} 
 & {Char.} & {Obj.} & {Char.} & {Obj.} & {Char. + Obj.} & {Char.} & {Obj.} & {Char. + Obj.} \\  
\hline
 Wan-VACE-14B & {5.54} & 4.74 & 2.76 & 3.96 & 5.14 & 3.58 & 3.48 & 4.28 & 4.10 \\
\hline
 \rowcolor{lightblue}
Wan2.2-5B + MetaCanvas  & \textbf{5.58} & \textbf{5.78} & \textbf{4.08} & \textbf{5.44} & \textbf{6.60} & \textbf{4.60} & \textbf{5.46} & \textbf{4.74} & \textbf{5.28}     \\ 
\hlineB{3}
\multirow{2}{*}{\textbf{Subject-Consistency}} & \multicolumn{2}{c}{{Single ID}} & \multicolumn{3}{c}{{Multiple IDs}} & \multicolumn{3}{c}{{Scene}} & \multirow{2}{*}{{Average$\uparrow$}}  \\ \cline{2-9} 
 & {Char.} & {Obj.} & {Char.} & {Obj.} & {Char. + Obj.} & {Char.} & {Obj.} & {Char. + Obj.} \\  
\hline
 Wan-VACE-14B & \textbf{8.86} & \textbf{8.96} & \textbf{4.52} & \textbf{7.02} & 6.66 & 4.54 & 4.90 & \textbf{5.26} & \textbf{6.34} \\
\hline
 \rowcolor{lightblue}
Wan2.2-5B + MetaCanvas  & 7.46 & {7.80} & 4.06 & {6.46} & \textbf{6.68} & \textbf{5.04} & \textbf{5.64} & 4.90 & {6.01}     \\ 
\hlineB{3}
\end{tabular}}
\end{center}
\end{table*}

\noindent\textbf{Evaluation strategy.}  
To assess the edited videos, we employ the following evaluation methods jointly:
\begin{itemize}[leftmargin=1em]
    \setlength{\itemsep}{0pt}
    \setlength{\parskip}{0pt}
    \setlength{\parsep}{0pt}
    \item \textbf{VBench~\citep{huang2024vbench++} score.} We use the quality scores from VBench to evaluate the overall video quality, which is orthogonal to how well the edited video follows the editing instructions. The quality score is computed as the average of subject consistency, background consistency, motion smoothness, aesthetic quality, dynamic degree, and imaging quality. We follow the VBench codebase to calculate these metrics.
    \item \textbf{MLLM evaluation.} We adopt a strategy similar to the GEdit~\citep{liu2025step1x} benchmark for video editing. Specifically, GPT-40~\citep{openai2024gpt4o} is used to evaluate semantic and quality scores, following the same prompt templates as GEdit but replacing ``input image'' with ``input frames''. To improve API efficiency, video frames are sampled at 1 FPS. The overall score is calculated in the same manner as in the original GEdit benchmark.
    \item \textbf{Human evaluation.} We conduct human evaluation on videos generated by \method{} and the two strongest baseline methods, Lucy-Edit-Dev-v1.1~\citep{decart2025lucyedit} and Ditto~\citep{bai2025scaling}. A total of 50 human evaluators participate, with videos randomly shuffled to ensure blind scoring. Evaluators are provided with the input video, editing prompt, and the three edited videos, and are asked to identify the videos with the best and worst editing accuracy, as well as the best and worst consistency. Editing accuracy measures prompt-following ability, similar to semantic evaluation in the MLLM method, while consistency measures whether the edited video is spatially and temporally aligned with the input video while maintaining subject and background consistency. The winning rates for each method are reported in~\Cref{table:results_video_editing}.
\end{itemize}

\section{In-Context Video Generation Task}
\label{sec:appendix_in_context_video_generation_task}

\subsection{Training Dataset Details}

Below, we provide a detailed illustration of our data curation strategy. Given the first frame of a video, we first use Qwen2.5-VL~\citep{Qwen2.5-VL} to generate editing prompts that extract the dominant objects and characters, as well as the background scene, from the image. We then provide the image along with these extraction prompts to Qwen-Image-Edit~\citep{wu2025qwen} to generate images containing the extracted characters, objects, and scenes as reference images. 

Additionally, we observed that when Qwen-Image-Edit extracts objects or characters, it often uses a white background, whereas the input reference images usually contain diverse backgrounds. To address this, we use Qwen2.5-VL to imagine suitable and natural backgrounds for the character and object reference images, and then use Qwen-Image-Edit to generate new edited images with these backgrounds. Using this process, we create a total of 70k in-context video data samples for training.

\subsection{Evaluation Benchmark Details}
As in-context video generation from reference images is a relatively new task and no well-established open-source evaluation benchmark currently exists, we adapt OmniContext~\citep{wu2025omnigen2}, an in-context image generation benchmark, into a video generation benchmark. We named this new benchmark as \textbf{OmniContext-Video}. Specifically, we reuse the same editing prompts and reference images from OmniContext without any modifications, but the objective is now to generate videos instead of images based on these reference images and user text instructions. This design allows us to directly leverage the original OmniContext evaluation protocol, by simply replacing the prompt instruction to evaluate the generated ``video'' instead of the generated ``image.'' For API efficiency, we sample the generated video at 1 FPS before providing it to GPT-4.1~\citep{openai2024gpt41} for evaluation. Consistent with the original benchmark, we evaluate both prompt-following and subject-consistency. \Cref{table:results_omnicontext_video} in the main paper reports the overall score averaged across these two dimensions, and we provide an expanded version including individual scores for each dimension in~\Cref{table:appendix_results_omnicontext_video} for completeness.

\section{Video Tasks Training Details}
\label{sec:appendix_video_tasks_training_details}

\begin{figure*}[t]
  \centering
  \includegraphics[width=0.99\linewidth]{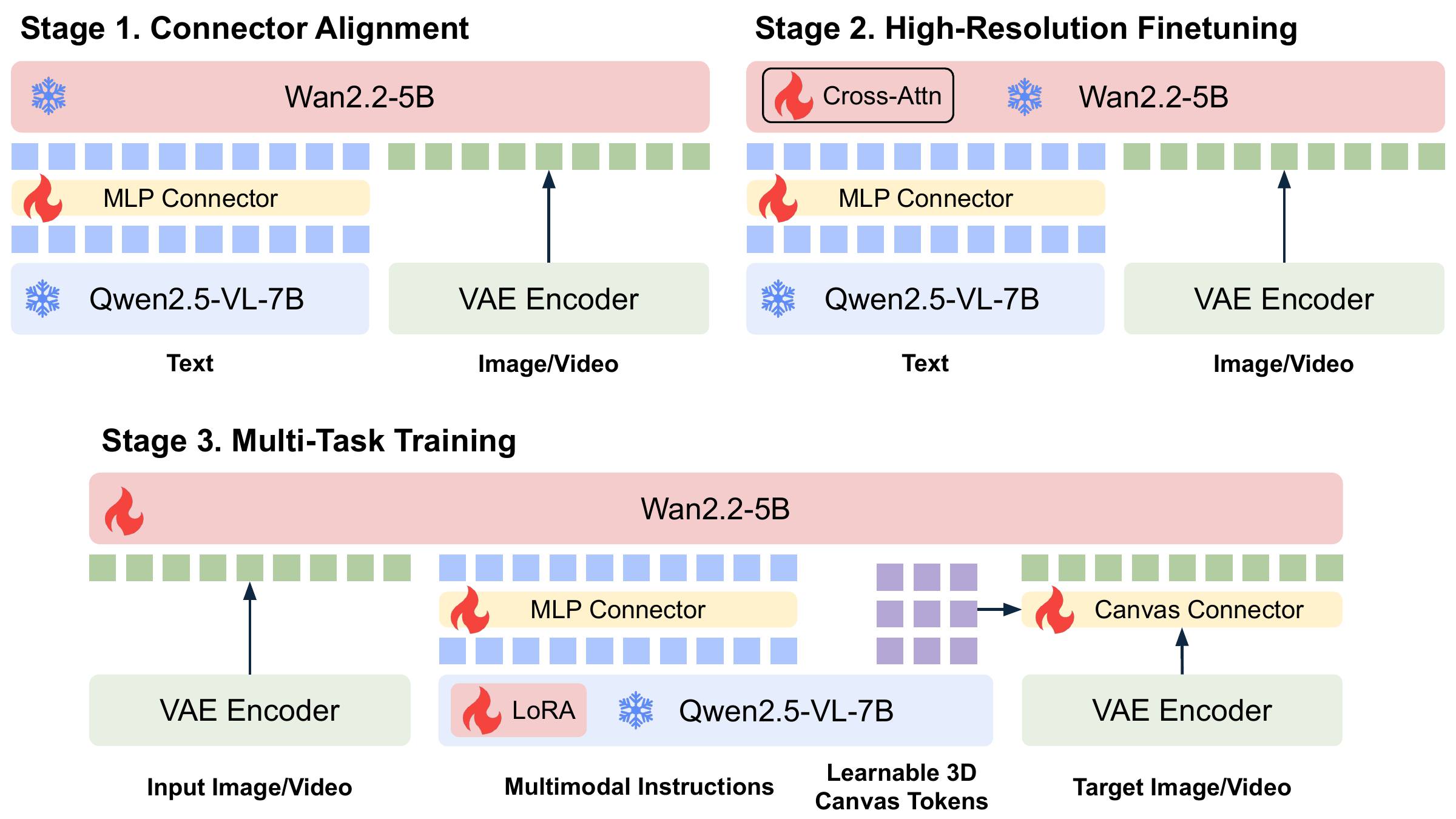}
  \caption{
  \textbf{\method implementation architecture details in~\Cref{subsec:results_video_generation_editing}.} We adopt Wan2.2-5B~\citep{wan2025wan} as the DiT and Qwen2.5-VL-7B~\citep{Qwen2.5-VL} as the MLLM. We illustrate the trainable components in each training stage respectively.
  }
  \vspace{-2mm}
\label{fig:appendix_video_training_stages} 
\end{figure*}

As detailed in~\Cref{subsec:exp_setup_training}, we adopt a three-stage training strategy for video tasks. To help readers better understand the trainable components in each stage, we visualize the training frameworks in~\Cref{fig:appendix_video_training_stages}. Additionally, we provide a comprehensive overview of the training hyperparameters, model parameters, and data sampling ratios in~\Cref{table:video_training_hyperparameters} to ensure reproducibility.

We note that the first two stages are designed solely to align the MLLM (i.e., Qwen2.5-VL-7B~\citep{Qwen2.5-VL}) with the video generation model (i.e., Wan2.2-5B~\citep{wan2025wan}), without involving any canvas tokens or canvas connector training. These stages are trained on basic tasks, including text-to-image generation, image reconstruction, and text-to-video generation. We allocate a relatively large compute budget to these stages, as sufficient training is crucial to properly align the MLLM with the diffusion model. Insufficient training in these stages would make it difficult for the model to capture fine-grained prompt information in stage 3 due to diluted learning on the the above basic tasks.

The third stage focuses on multi-task training, incorporating our model-specific canvas components. Compared with the first two stages, stage 3 requires relatively fewer training steps, completing in only 10k steps. We also observe that multi-task training provides mutual benefits across tasks. For example, in~\Cref{fig:visualization_t2v}, \method{} demonstrates improved understanding of styles, an ability likely learned from the image and video style editing data.

\begin{table*}[h!]
\begin{center}
\setlength{\tabcolsep}{0.9em}
\renewcommand{\arraystretch}{1.0}
\caption{\small{\textbf{Training hyperparameters on Wan2.2-TI2V-5B with MetaCanvas across different stages.}}}
\label{table:video_training_hyperparameters}
\vspace{-1mm}
\scalebox{0.95}{
\begin{tabular}{lccc}
\hlineB{3}
\multirow{3}{*}{\textbf{Hyperparameters}} & \multicolumn{3}{c}{\textbf{Training Stages}}  \\ \cline{2-4} 
 & \textbf{Stage 1} & \textbf{Stage 2} & \textbf{Stage 3}  \\
 & Connector Alignment & High Resolution Finetuning & Multi-task Training \\ \hline
\multicolumn{2}{l}{\textcolor{gray}{Training Hyperparameters}} \\
Learning Rate & 1e-3 & 2e-5 & 1e-5  \\
LR Scheduler & Constant & Constant & Constant  \\
Weight Decay & 0.0 & 0.0 & 0.0 \\
Gradient Clip & 1.0 & 1.0 & 1.0  \\
Optimizer & AdamW & AdamW & AdamW \\
Warm-Up Steps & 100 & 100 & 100  \\
Training Steps & 100k & 15k & 10k  \\
Image/Video Resolution & 480$\times$832 & 704$\times$1280 & 704$\times$1280  \\
Video \# Frames & - & 121 & 121  \\
Video FPS & - & 24 & 24  \\
Image Batch Size / GPU & 32 & 32 & 16  \\
Video Batch Size / GPU & - & 3 & 1  \\
\# A100 GPUs & 128 & 128 & 128  \\
\hline
\multicolumn{2}{l}{\textcolor{gray}{Model Parameters}}
\\
Trainable Modules & MLP Connector &  Connector + Cross-Attn & All Params  \\
MLLM Tokens Per Image & 144 & 144 & 144  \\
MLLM LoRA Rank & - & - & 64  \\
\# Canvas Tokens & - & - & 11$\times$20$\times$3$=$660  \\
\hline
\multicolumn{2}{l}{\textcolor{gray}{Data Sampling Ratio}} \\
Text-to-Image & 80\% & 16\% & 5\%  \\
Image Reconstruction & 20\% & 4\% & - \\
Image Editing & - & - & 15\%  \\
Text-to-Video & - & 80\% & 15\% \\
Image-to-Video & - & - & 15\%  \\
Video Editing & - & - & 25\%  \\
Reference-image-to-Video & - & - & 25\%  \\
\hlineB{3}
\end{tabular}}
\end{center}
\vspace{-4mm}
\end{table*}

\section{Visualizations}
\label{sec:appendix_visualizations}

In this section, we first illustrate how we plot the canvas map, and then provide additional visualization examples for different tasks.

\begin{figure*}[h!]
  \centering
  \includegraphics[width=0.94\linewidth]{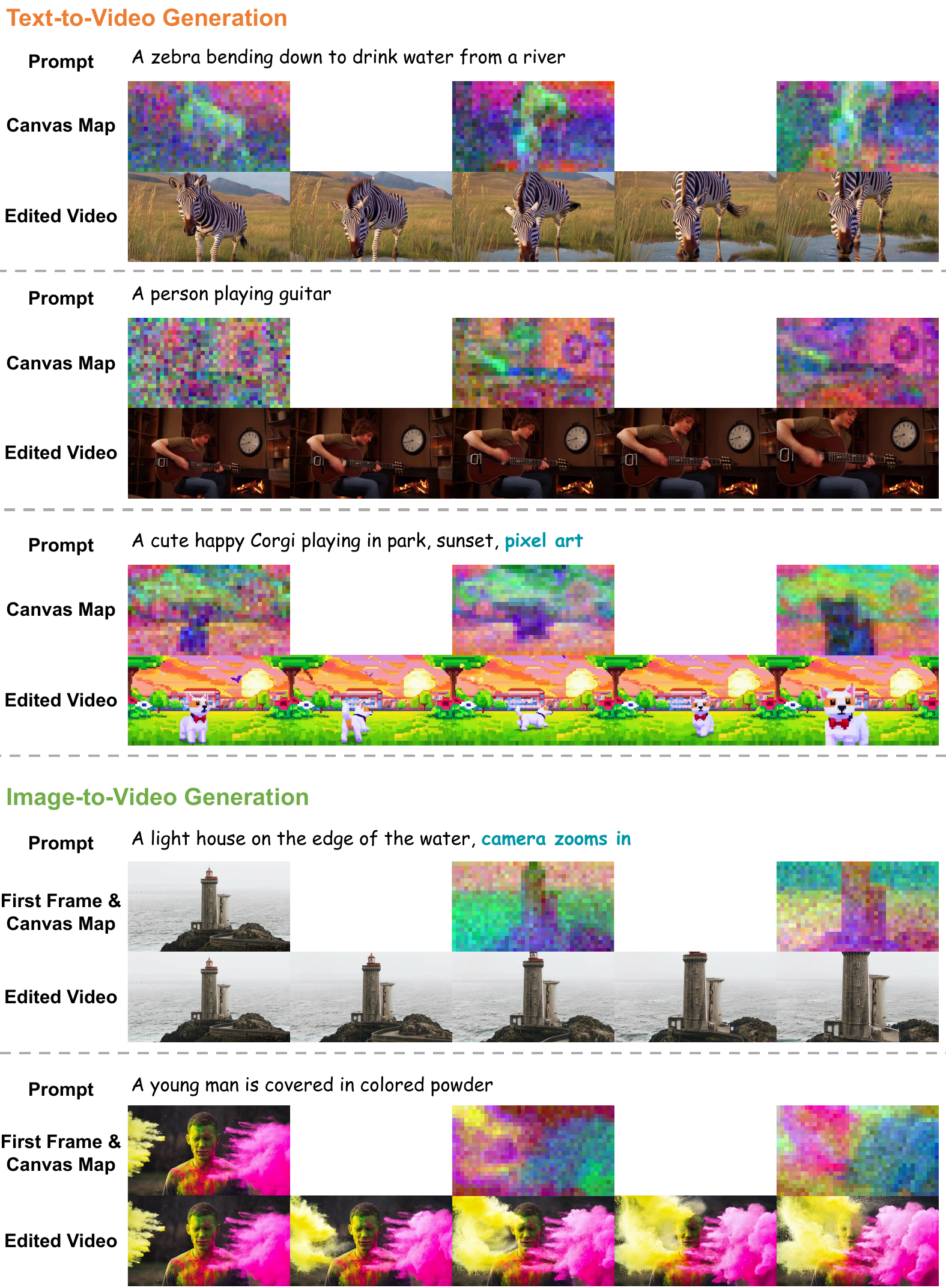}
  \vspace{-1mm}
  \caption{
  \textbf{Visualization of the {\color{cyan} PCA feature maps of the 3 canvas keyframes} for the {\color{orange}text-to-video generation} and {\color{ForestGreen}image-to-video generation} tasks.}
  Note that for the image-to-video generation task, because the first frame is assigned $t = 0$ during inference with Wan2.2-5B, we omit the visualization for the first canvas keyframe.
  Interpretation of the canvas map visualization is discussed in~\Cref{sec:appendix_visualizations}.} 
\label{fig:visualization_canvas_map_t2v_i2v} 
\end{figure*}

\begin{figure*}[h!]
  \centering
  \includegraphics[width=0.95\linewidth]{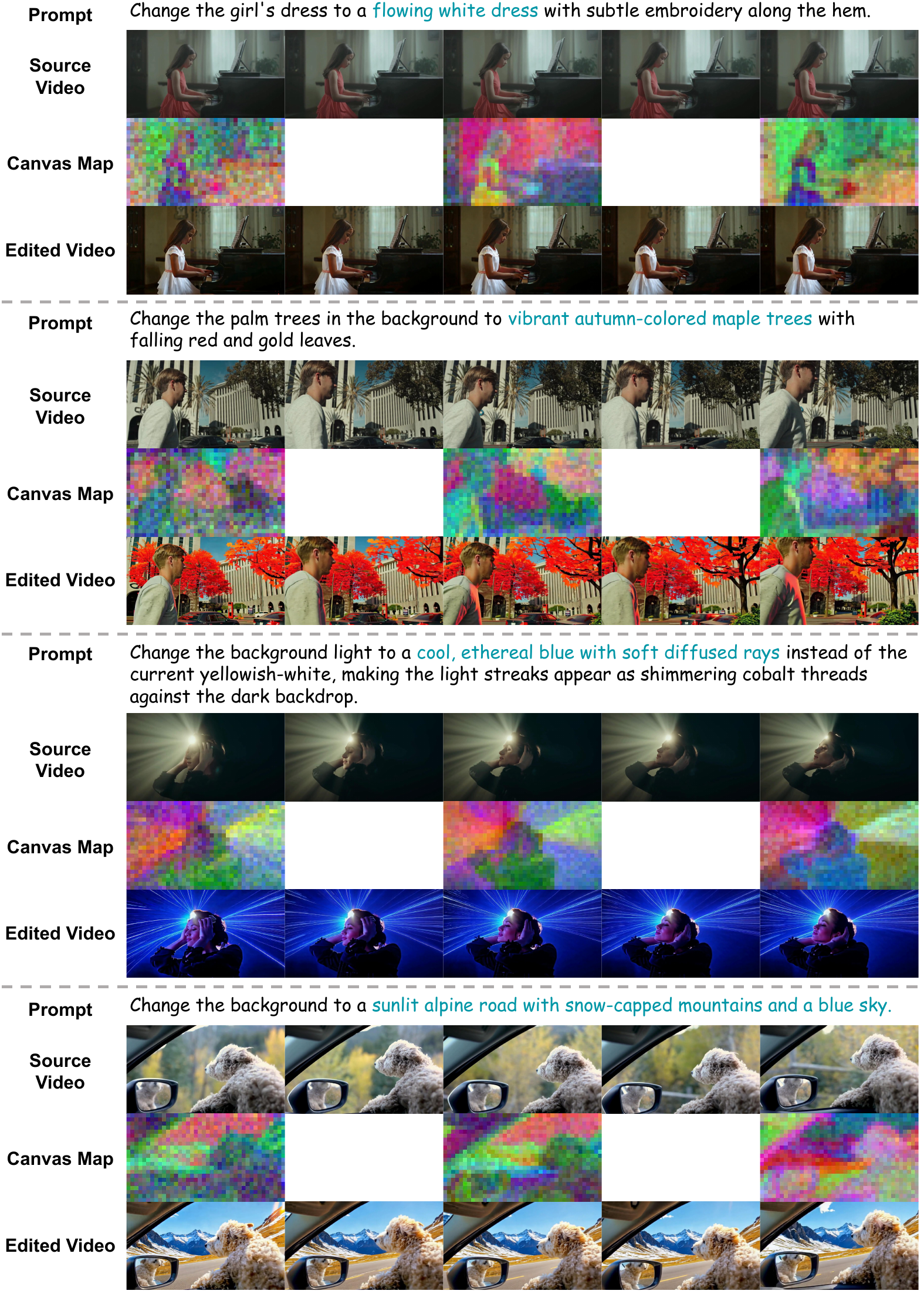}
  \vspace{-1mm}
  \caption{
  \textbf{Visualization of the {\color{cyan} PCA feature maps of the 3 canvas keyframes} for the {\color{purple}video editing} task.} Interpretation of the canvas map visualization is discussed in~\Cref{sec:appendix_visualizations}.} 
\label{fig:visualization_canvas_map_v2v} 
\end{figure*}

\noindent\textbf{Canvas map visualization.} 
Following~\citep{tumanyan2023plug}, we apply PCA to the features produced by the \method{} connector. The features are extracted at approximately the middle of the diffusion timesteps (e.g., $t=499$ for the text-to-image generation task in~\Cref{fig:attention_map_visualization}, and $t=522$ for the video tasks). For each extracted feature, we apply PCA and visualize the top three principal components. Note that the colors on the PCA map only represent salient subjects or backgrounds, and the color assigned to a given subject or background may vary across different canvas keyframes.
As a supplement to~\Cref{fig:attention_map_visualization}, we provide additional visualizations of the canvas maps for text-to-video generation and image-to-video generation tasks in~\Cref{fig:visualization_canvas_map_t2v_i2v}, and visualizations for video tasks in~\Cref{fig:visualization_canvas_map_v2v}.

\begin{figure*}[h!]
  \centering
  \includegraphics[width=0.95\linewidth]{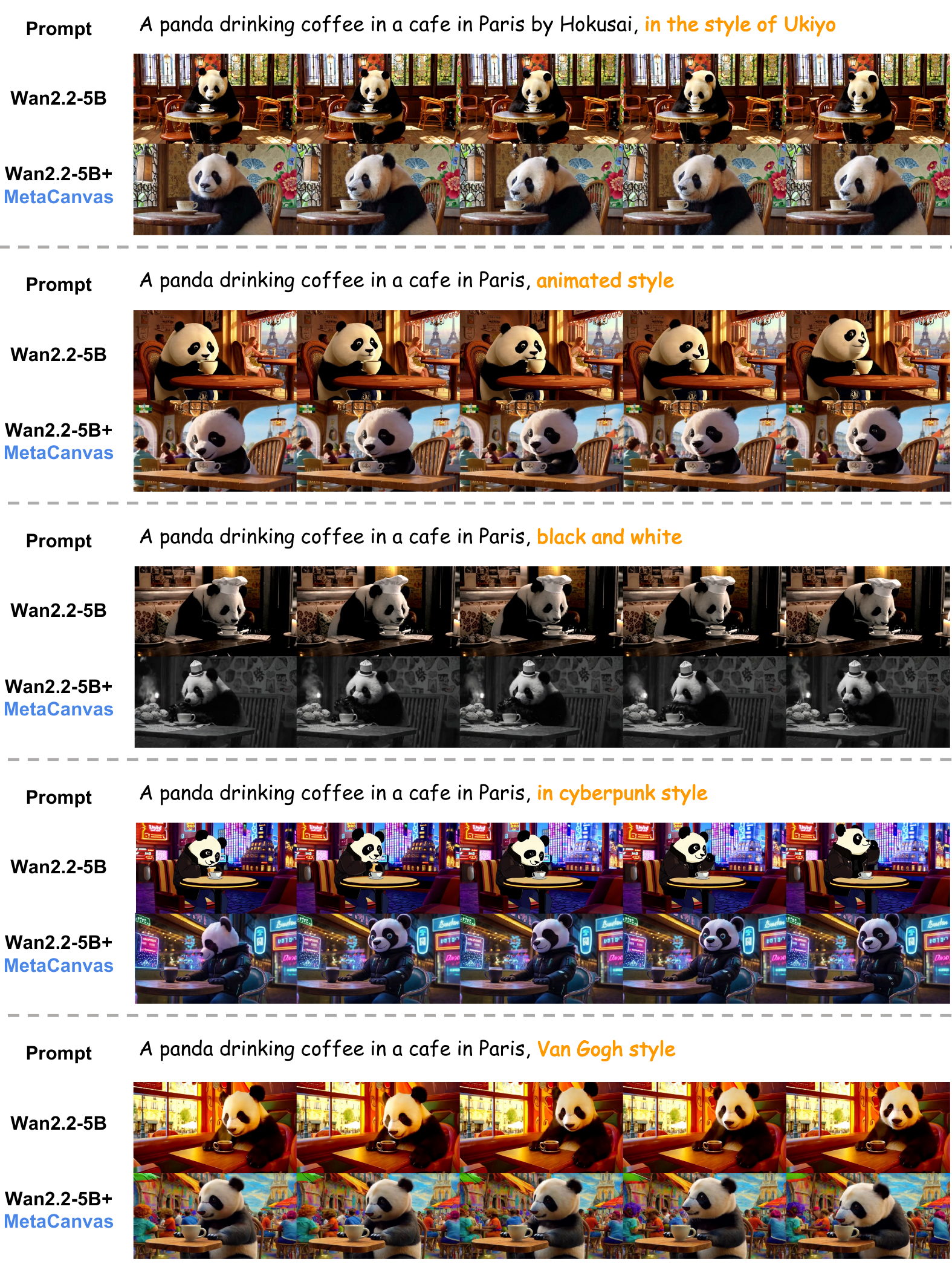}
  \vspace{-2mm}
  \caption{
  \textbf{Visualization on {\color{orange}text-to-video generation}  task.} The input prompts are sourced from VBench~\citep{huang2023vbench}. Multi-task training in Stage 3 enables \method{} to better understand stylistic variations in prompts.
  }
\label{fig:visualization_t2v} 
\end{figure*}

\noindent\textbf{Visualizations for T2V generation task.} 
In~\Cref{fig:visualization_t2v}, we present visualizations for the text-to-video generation task, specifically on the ``appearance style'' category in VBench~\citep{huang2023vbench}. As our model is jointly trained on diverse tasks, including video global style editing, we observe that such joint training enables \method{} to better understand stylistic variations in prompts.

\begin{figure*}[h!]
  \centering
  \includegraphics[width=0.95\linewidth]{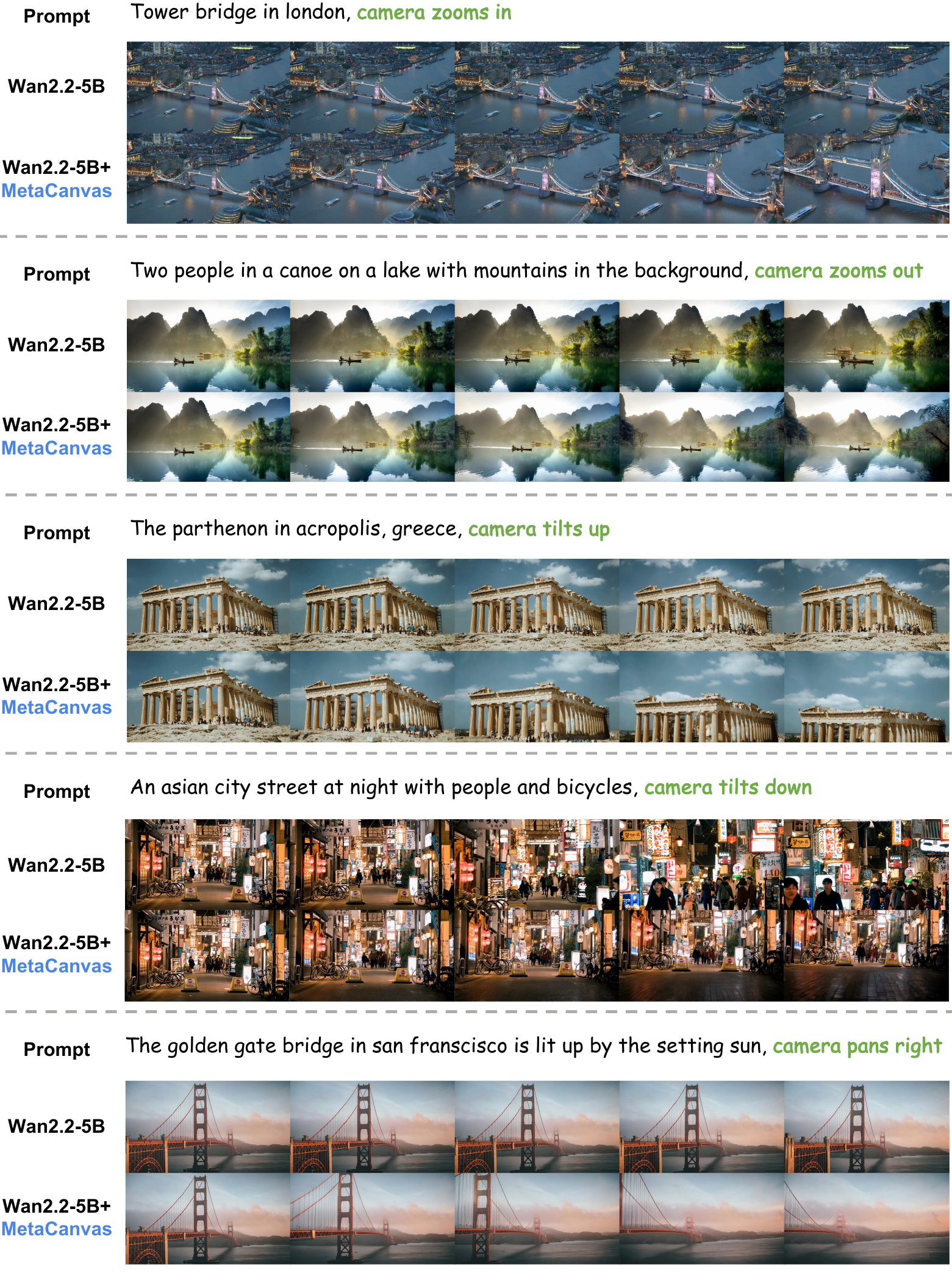}
  \vspace{-2mm}
  \caption{
  \textbf{Visualization on {\color{ForestGreen}image-to-video generation} task.} The input images and prompts are sourced from VBench~\citep{huang2024vbench++}. Compared with the base model, \method{} better understands instructions involving camera motions.
  }
\label{fig:visualization_i2v} 
\end{figure*}

\noindent\textbf{Visualizations for I2V generation task.} 
In~\Cref{fig:visualization_i2v}, we show visualizations for the image-to-video generation task, focusing on the ``camera motion'' category in VBench~\citep{huang2024vbench++}. Compared with the base Wan2.2-5B model, \method{} more effectively interprets instructions involving camera motions. In~\Cref{table:results_vbench_i2v}, we further provide the expanded results corresponding to~\Cref{table:results_t2v_i2v} in the main paper.

\begin{figure*}[h!]
  \centering
  \includegraphics[width=0.95\linewidth]{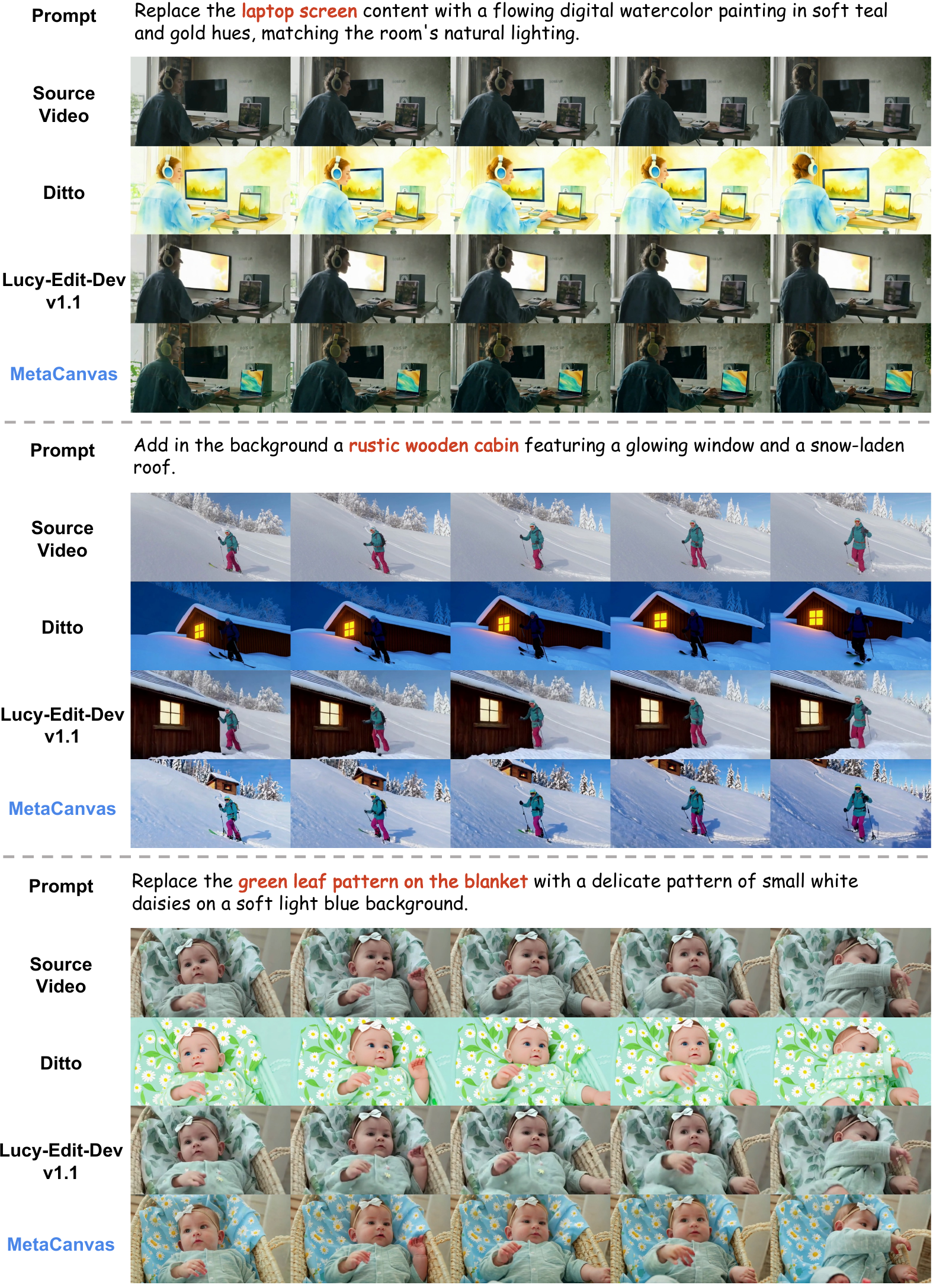}
  \vspace{-2mm}
  \caption{
  \textbf{Visualization for the {\color{purple} video local editing} task (part 1/3).} \method{} achieves more precise grounding of the target objects.
  }
\label{fig:visualization_v2v_p1} 
\end{figure*}

\begin{figure*}[h!]
  \centering
  \includegraphics[width=0.95\linewidth]{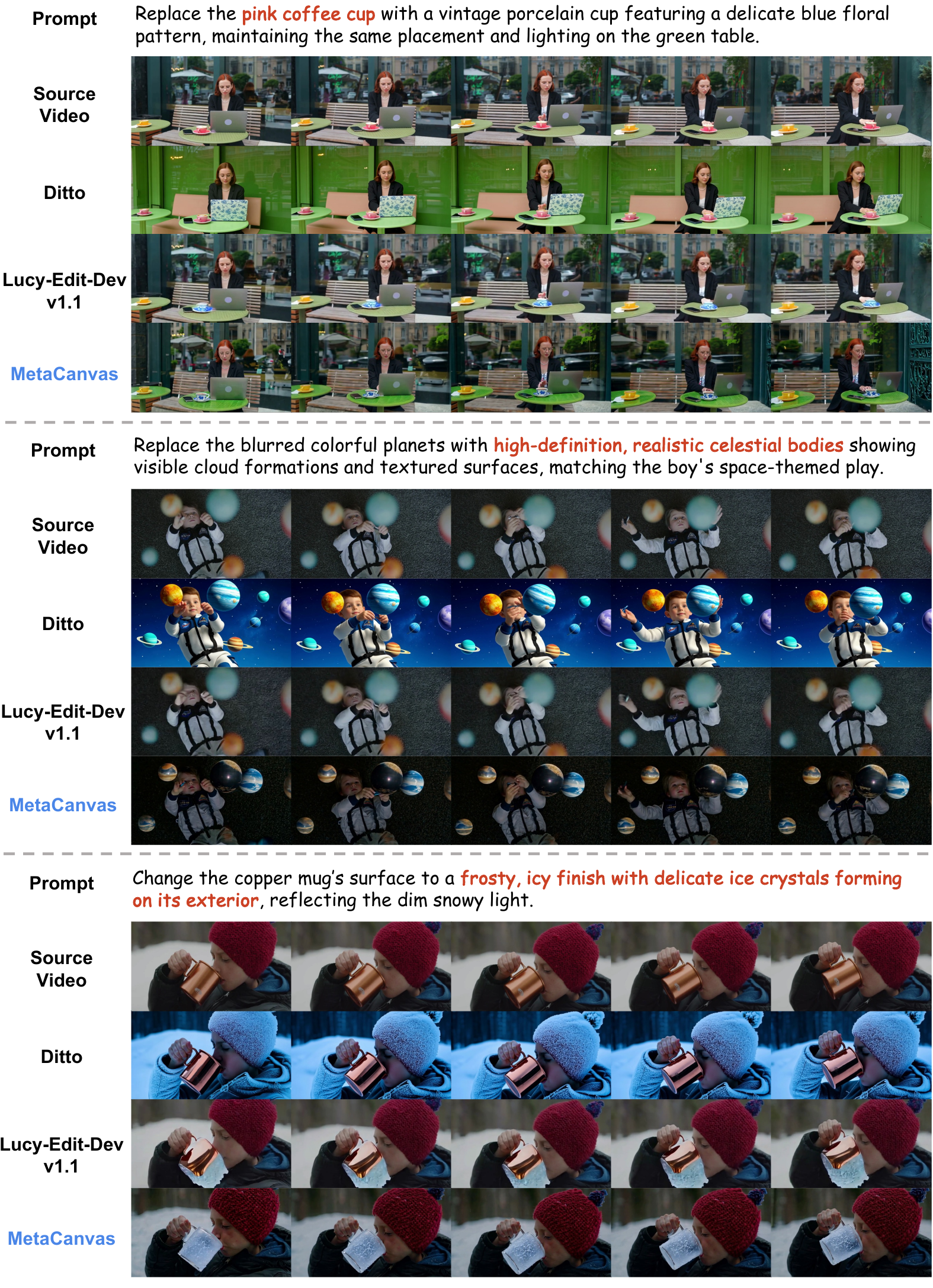}
  \vspace{-2mm}
  \caption{
  \textbf{Visualization for the {\color{purple}video local editing} task (part 2/3).} \method{} achieves more precise grounding of the target objects.
  }
\label{fig:visualization_v2v_p2} 
\end{figure*}

\begin{figure*}[h!]
  \centering
  \includegraphics[width=0.95\linewidth]{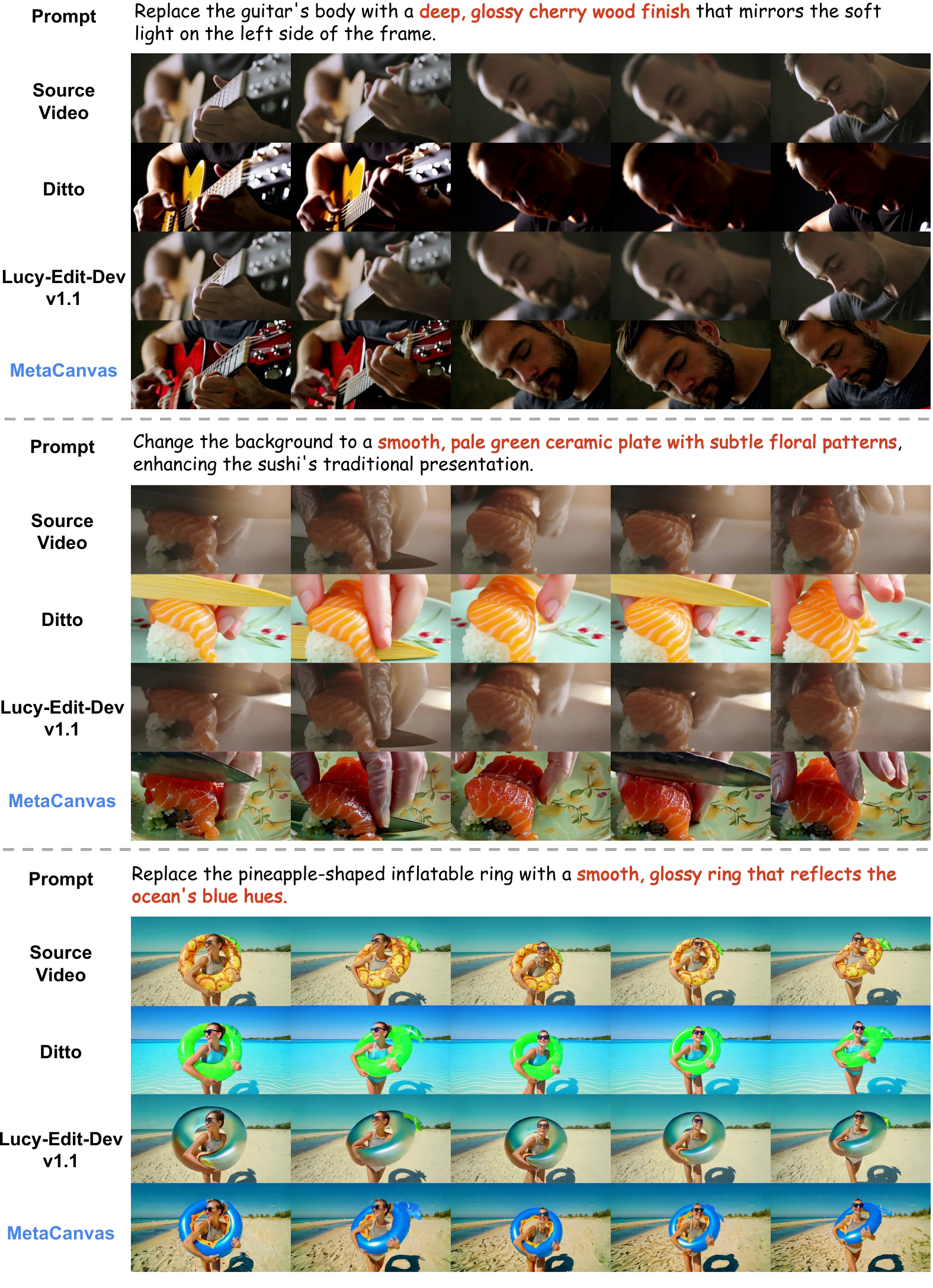}
  \vspace{-2mm}
  \caption{
  \textbf{Visualization for the {\color{purple}video local editing} task (part 3/3).} \method{} achieves more precise grounding of the target objects.
  }
\label{fig:visualization_v2v_p3} 
\end{figure*}

\begin{figure*}[h!]
  \centering
  \includegraphics[width=0.95\linewidth]{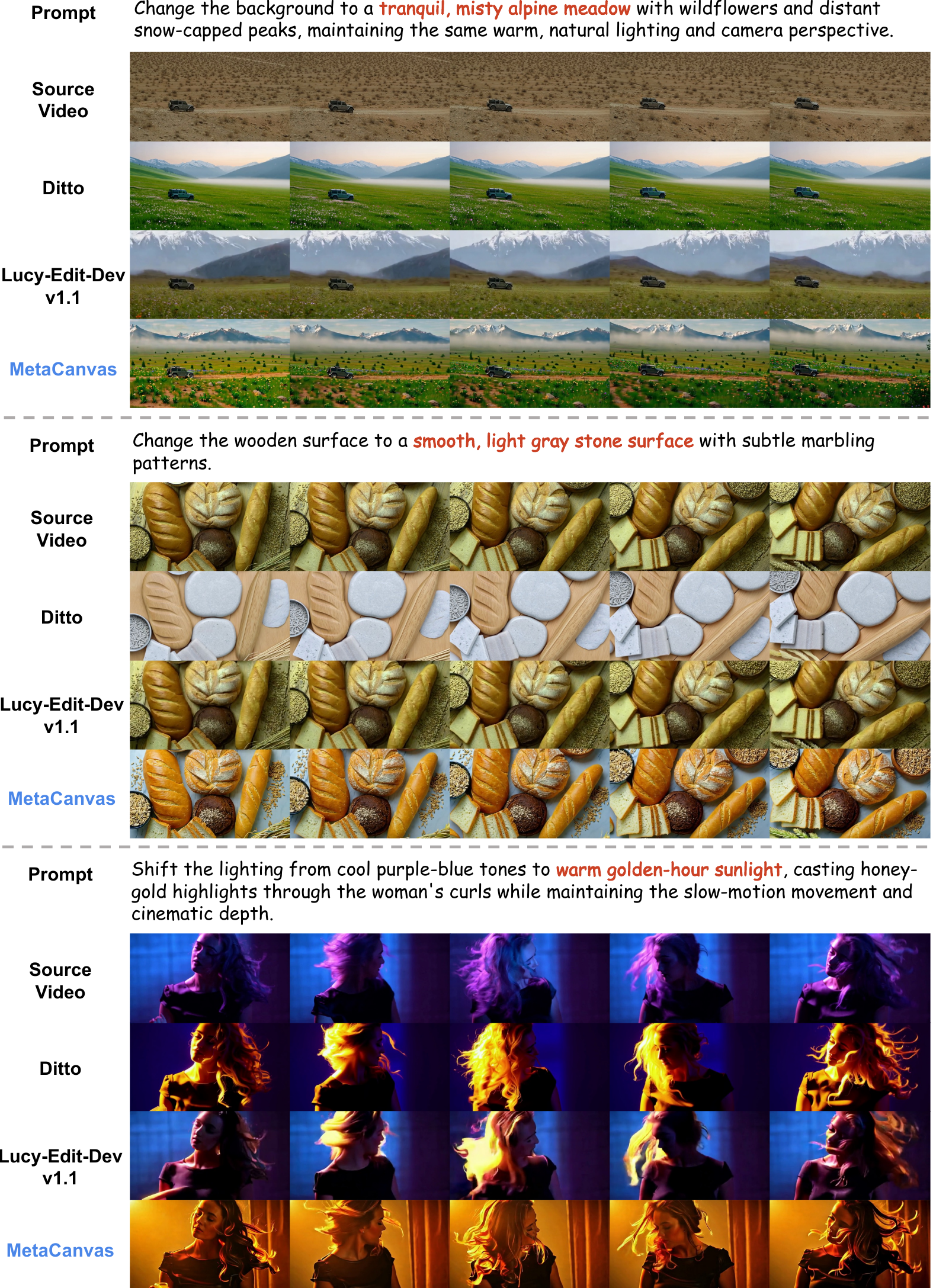}
  \vspace{-1mm}
  \caption{
  \textbf{Visualization for the {\color{purple} video background editing} task.} \method{} achieves more precise instruction following and grounding of the target background.
  }
\label{fig:visualization_v2v_background} 
\end{figure*}

\begin{figure*}[h!]
  \centering
  \includegraphics[width=0.95\linewidth]{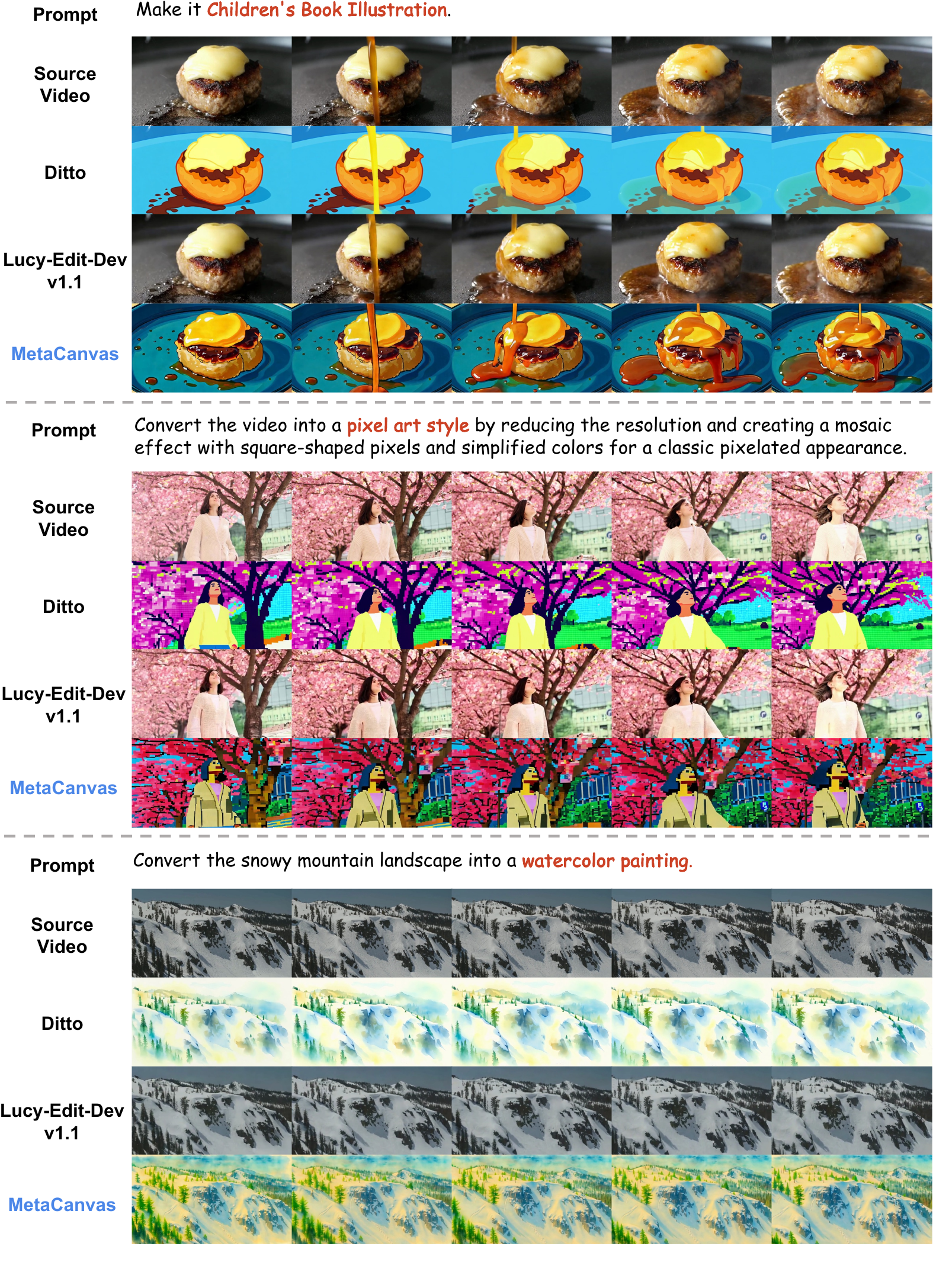}
  \vspace{-8mm}
  \caption{
  \textbf{Visualization for the {\color{purple} video global style editing} task.} \method{} achieves results comparable to Ditto~\citep{bai2025scaling}, whereas Lucy-Edit-Dev-v1.1~\citep{decart2025lucyedit} fails on this task.
  }
\label{fig:visualization_v2v_style} 
\end{figure*}

\noindent\textbf{Visualizations for video editing task.} 
In~\Cref{fig:visualization_v2v_p1,fig:visualization_v2v_p2,fig:visualization_v2v_p3}, we show visualizations for the (local) video editing task. Compared with other methods~\citep{bai2025scaling, decart2025lucyedit}, \method{} achieves more precise grounding of the target objects.
In~\Cref{fig:visualization_v2v_background}, we show visualizations for the video background editing task. Compared with other methods~\citep{bai2025scaling, decart2025lucyedit}, \method{} achieves more precise instruction following and grounding of the target background.
In~\Cref{fig:visualization_v2v_style}, we show visualizations for the video global style editing task. \method{} achieves results comparable to Ditto~\citep{bai2025scaling}, whereas Lucy-Edit-Dec-v1.1~\citep{decart2025lucyedit} fails on this task.

\begin{figure*}[h!]
  \centering
  \includegraphics[width=0.95\linewidth]{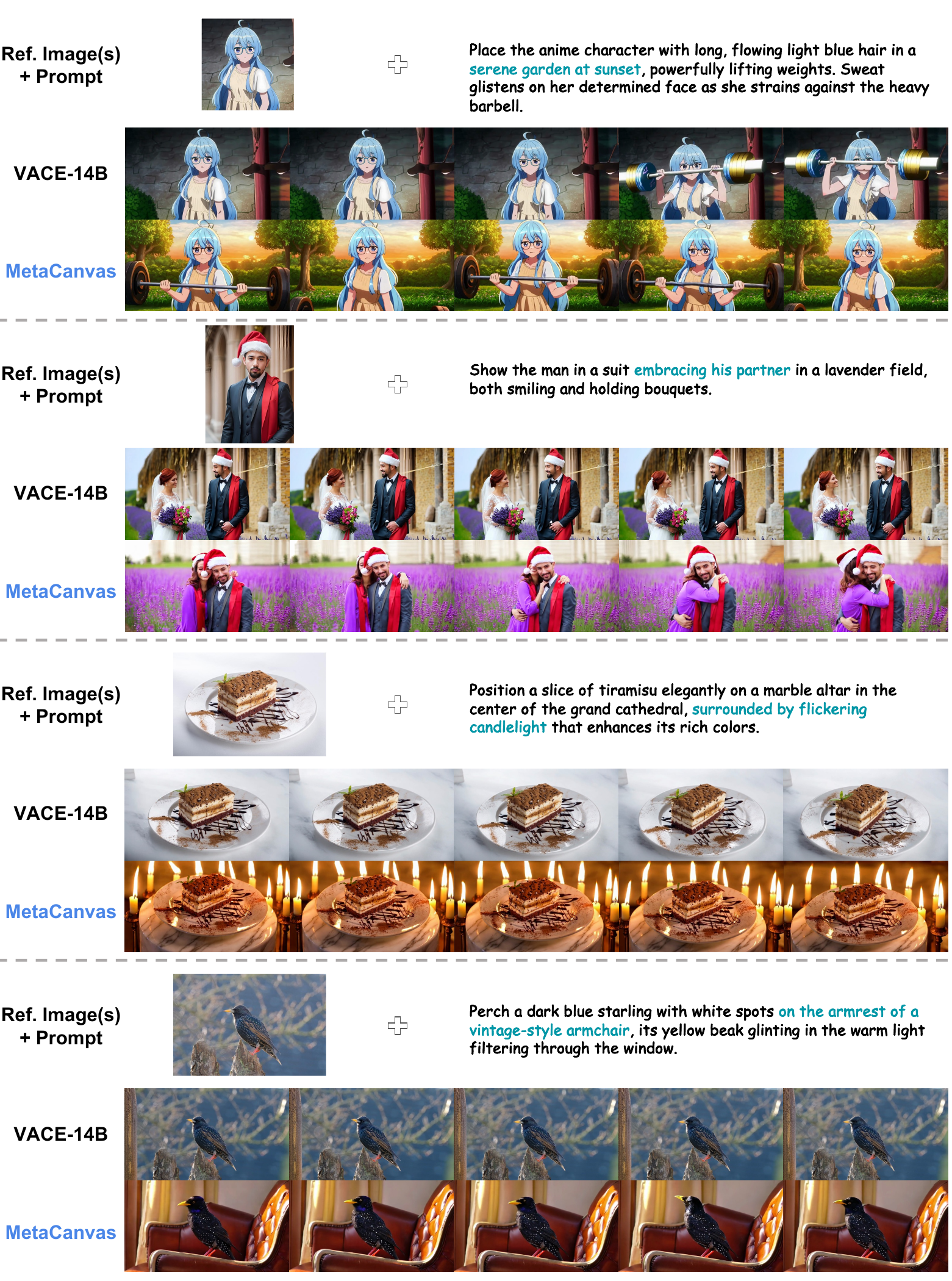}
  \vspace{-1mm}
  \caption{
  \textbf{Visualization for the {\color{RoyalBlue} in-context video generation} task with \textbf{single} character or object.} \method{} demonstrates improved prompt understanding and more accurately places the specified character or object in the appropriate scene, whereas VACE~\citep{jiang2025vace} incorrectly adheres too closely to the background of the reference image.
  }
\label{fig:visualization_ri2v_single} 
\end{figure*}

\begin{figure*}[h!]
  \centering
  \includegraphics[width=0.95\linewidth]{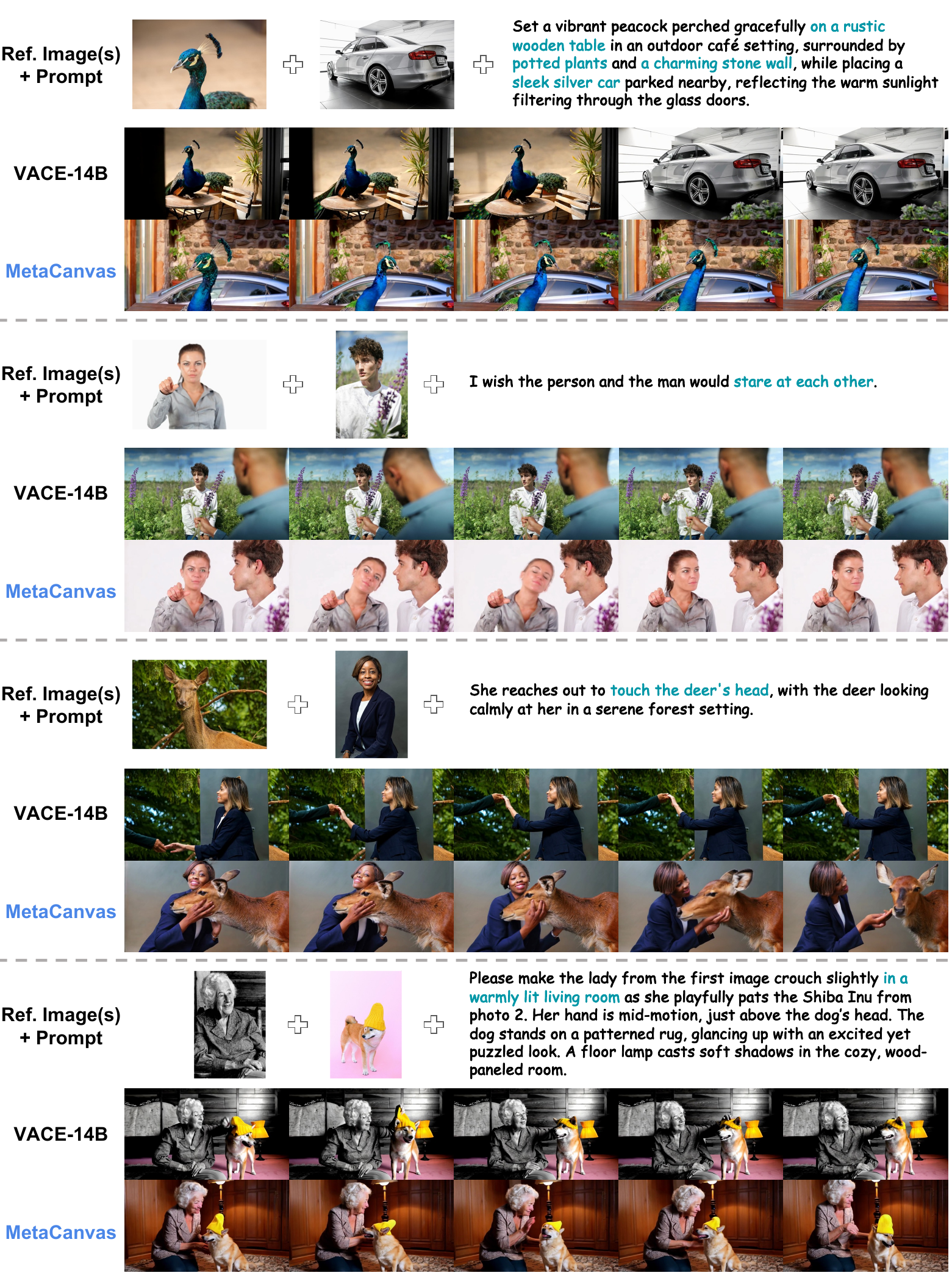}
  \vspace{-1mm}
  \caption{
  \textbf{Visualization for the {\color{RoyalBlue} in-context video generation} task with \textbf{multiple} characters and/or objects.} \method{} demonstrates an improved ability to naturally compose the reference images into the appropriate scene, whereas VACE~\citep{jiang2025vace} either adheres too closely to the background of the reference images or fails to compose both reference images correctly.
  }
\label{fig:visualization_ri2v_multiple} 
\end{figure*}

\begin{figure*}[h!]
  \centering
  \includegraphics[width=0.95\linewidth]{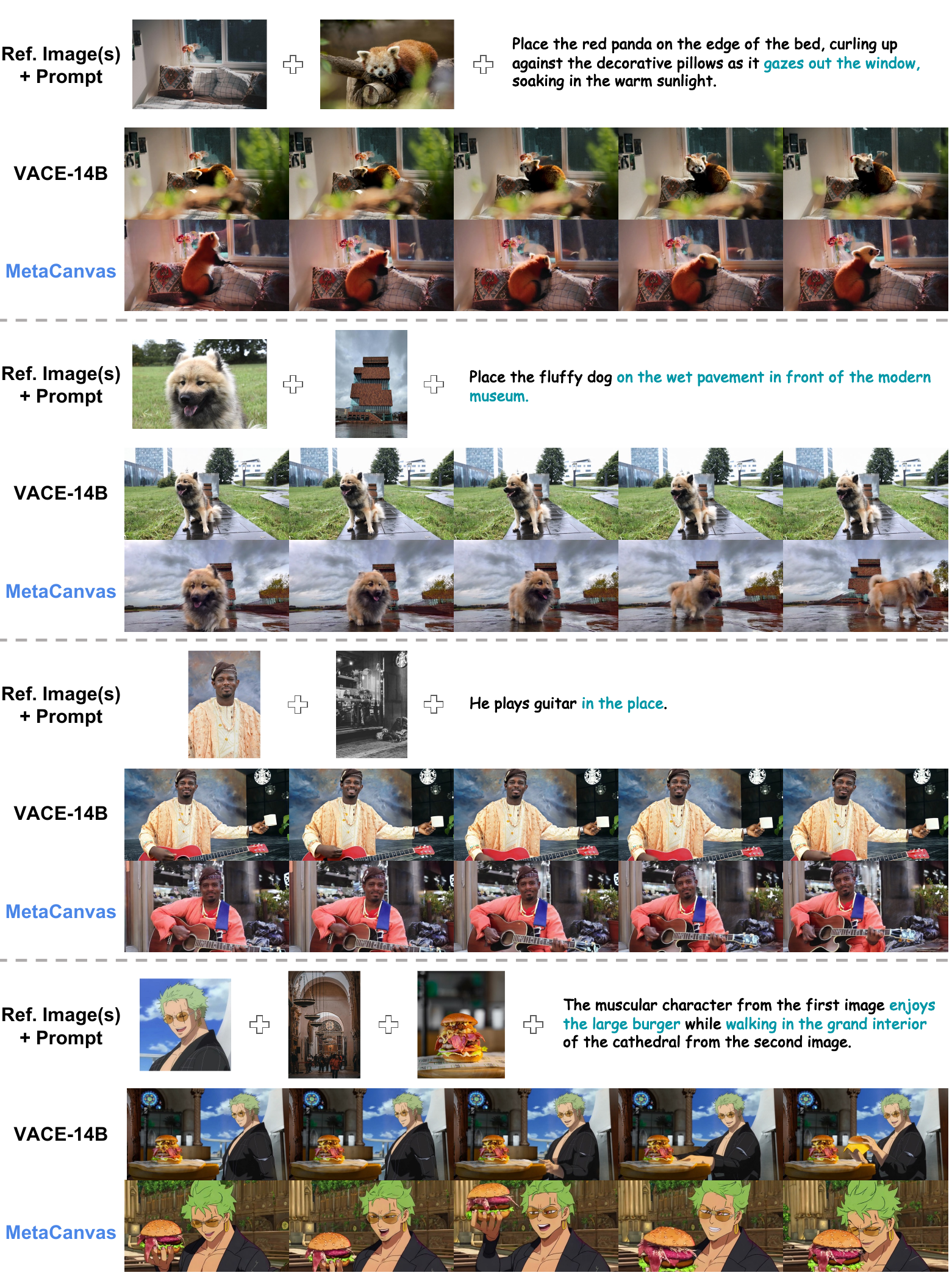}
  \vspace{-1mm}
  \caption{
  \textbf{Visualization for the {\color{RoyalBlue} in-context video generation} task with character and/or object in a \textbf{scene}.} \method{} demonstrates improved prompt understanding and naturally composes the reference images into the appropriate scene. However, both methods produce imperfect videos when the number of reference images increases to three.
  }
\label{fig:visualization_ri2v_scene} 
\end{figure*}

\noindent\textbf{Visualizations for in-context video generation task.} 
In~\Cref{fig:visualization_ri2v_single,fig:visualization_ri2v_multiple,fig:visualization_ri2v_scene}, we present visualizations for the in-context video generation task using reference images. We observe that the comparison baseline (i.e., VACE-14B~\citep{jiang2025vace}) suffers from a “copy–paste’’ effect, where the background of the reference image is replicated in the generated video. This makes it difficult for the model to follow the instructions and to naturally compose multiple reference images into the correct background. In addition, both VACE-14B and \method{} produce imperfect videos when the number of reference images increases to three. We hypothesize that an improved data creation strategy with more and higher-quality training data, and assigns a higher proportion of examples to multi-reference video generation, could help alleviate this issue.

\section{Simplified Code Implementation}

Our codebase builds upon the publicly available BLIP3o codebase~\citep{chen2025blip3ofamilyfullyopen}. Detailed training procedures and model hyperparameters are provided in~\Cref{subsec:exp_setup_training} and~\Cref{table:t2i_training_hyperparameters}. We will release our code upon acceptance. For better reproducibility, we also provide a simplified PyTorch implementation of the DiT block in the canvas connector in~\Cref{alg:canvas_connector}.

\section{Limitations and Future Works}

In this work, we primarily focus on investigating the effectiveness of information transfer between MLLMs and diffusion Transformers via \method{}. Our approach follows prior work that bridges MLLMs with diffusion models through lightweight connector interfaces. One potential limitation of the current setup is that visual information (e.g., images and videos) is provided to both the MLLM and the diffusion models, following previous works~\citep{wei2025univideo, wu2025qwen, lin2025uniworld, wu2025omnigen2, liu2025step1x} to maximize performance. It would be interesting to explore whether a more elegant framework could be designed that passes all visual information solely to the MLLM, allowing DiT to directly render images and videos without repeated conditioning on visual inputs.

Additionally, we evaluated the effectiveness of \method{} across diverse tasks using text, image, video, or combinations thereof as inputs. However, we note that the quality of our curated training data is not optimal, and the scale of the data is limited for some of the tasks. For instance, we observed that the success rate for in-context video generation from three or more reference images is not high. Expanding the task-specific dataset could further improve performance.

\clearpage

\begin{center}
\begin{minipage}{0.99\textwidth}
\begin{algorithm}[H]
\caption{Simplified PyTorch Implementation for the \textbf{DiT Block in Canvas Connector}}
\begin{verbatim}
import torch 
from torch import nn
import torch.nn.functional as F
from einops import rearrange
from diffusers.models.normalization import AdaLayerNormSingle
from diffusers.models.transformers.sana_transformer import SanaTransformerBlock

class CanvasConnectorDiTBlock(nn.Module):

    def __init__(self, in_channels=48, embed_dim=3072, num_layers = 1):
        super().__init__()
        self.embed_dim = embed_dim
        self.time_embed = AdaLayerNormSingle(embed_dim)
        self.patch_embed = nn.Conv3d(in_channels, embed_dim,
            kernel_size=(1, 2, 2), stride=(1, 2, 2)
        )
        self.out_proj = zero_module(nn.Linear(embed_dim, embed_dim))
        self.canvas_dit_blocks = nn.ModuleList(
            [
                SanaTransformerBlock(
                    dim=embed_dim,
                    num_attention_heads=embed_dim//32, 
                    attention_head_dim=32, 
                    num_cross_attention_heads=None, 
                    cross_attention_head_dim=None, 
                    cross_attention_dim=None,  
                    attention_bias=False, 
                )
                for _ in range(num_layers)
            ]
        )

    def forward(self, latent, interpolated_canvas_keyframes, timestep):
        latent_embed = self.patch_embed(latent)  
        b, t, _, h, w = latent_embed.shape

        # add after patchification
        hidden_states = latent_embed + interpolated_canvas_keyframes 
        hidden_states = rearrange(hidden_states, "b c t h w -> (b t) (h w) c") 

        timestep = timestep if timestep.ndim == 1 else timestep[:,-1]
        timestep, _ = self.time_embed(timestep).repeat(t, 1)

        # fused through the DiT block conditioned on the timestep
        for index_block, block in enumerate(self.canvas_dit_blocks):
            hidden_states = block(
                hidden_states, None, None, None, timestep, h, w, None, None
            )

        fused = self.out_proj(hidden_states) # linear projection before output
        fused = rearrange(fused, "(b t) (h w) c -> b c t h h w", b=b, h=h, w=w)
        latent_embed = latent_embed + fused # latent_embed is passed to the DiT
        return latent_embed
\end{verbatim}
\label{alg:canvas_connector}
\end{algorithm}
\end{minipage}
\end{center}

\twocolumn

\end{document}